\documentclass[11pt]{article}

\usepackage[preprint]{acl}

\usepackage{times}
\usepackage{latexsym}

\usepackage[T1]{fontenc}

\usepackage[utf8]{inputenc}

\usepackage{microtype}
\usepackage{enumitem}

\usepackage{inconsolata}

\usepackage{graphicx}

\usepackage{amsmath}
\usepackage{amssymb}
\usepackage{pifont}
\usepackage[table]{xcolor}
\usepackage{multirow}
\usepackage{wrapfig}
\usepackage{cleveref}
\newcommand{\xmark}{\ding{55}}%
\usepackage{marvosym}

 \usepackage{comment}
\usepackage{booktabs}
\usepackage{xcolor}
\usepackage{xspace}
\newcommand{\eg}{e.g.\@\xspace}

\newcommand{\cmark}{\ding{51}}%
\newcommand{\xxmark}{\textcolor[rgb]{ .819,  .819,  .819}{\xmark}}

\title{SAM3-I: Segment Anything with Instructions}

\author{\vspace{-0.3cm} \\
  {Jingjing Li\textsuperscript{1,$\dagger$}}, 
  {Yue Feng\textsuperscript{2,$\dagger$}}, 
  {Yuchen Guo\textsuperscript{3,$\dagger$}}, 
  {Jincai Huang\textsuperscript{4,$\dagger$}}, 
  {Wei Ji\textsuperscript{5,2,$\ddagger$,*}}, 
\\
  {Qi Bi\textsuperscript{6}}, 
  {Yongri Piao\textsuperscript{7}}, 
  {Miao Zhang\textsuperscript{7}}, 
  {Xiaoqi Zhao\textsuperscript{5}}, 
\\
  {Qiang Chen\textsuperscript{2}}, 
  {Shihao Zou\textsuperscript{8,*}}, 
  {Huchuan Lu\textsuperscript{7}}, 
  {Li Cheng\textsuperscript{1}} \vspace{.11cm}
\\
{\normalsize
  \textsuperscript{1}University of Alberta
  \textsuperscript{2}Tencent WeChat
  \textsuperscript{3}Northwestern University
  \textsuperscript{4}SUSTech
  \textsuperscript{5}Yale University
  }
\\
{\normalsize
  \textsuperscript{6}Utrecht University
  \textsuperscript{7}Dalian University of Technology
  \textsuperscript{8}SIAT, Chinese Academy of Sciences
}
\\
  \small{
    \textbf{$^\dagger$Equal Contribution},
    \textbf{$^\ddagger$Project Lead},
    \textbf{$^*$Corresponding Author}
  }
}

\begin{document}
\maketitle

\let\thefootnote\relax\footnotetext{\hspace{-.62cm} \textbf{Contact Emails}: jingjin1@ualberta.ca; wei.ji@yale.edu; {\small\{fengyue5717,ethan.chen1988\}}@gmail.com;sh.zou@siat.ac.cn.}

\begin{figure*}[t]
    \centering
    \vspace{-0.5cm}
    \includegraphics[width=\textwidth]{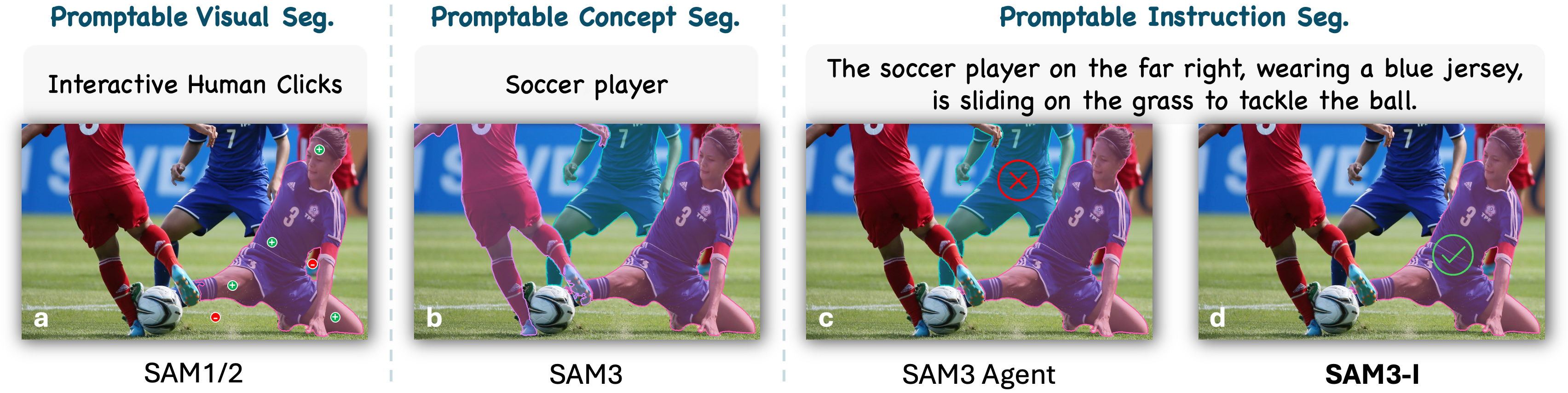}
    \vspace{-7.9mm}
    \caption{
    \textbf{Evolution of promptable segmentation in the SAM family}. (a) {P}romptable {V}isual {S}egmentation (\textbf{PVS}) in SAM1/2 relies on interactive visual prompts, with each prompt segmenting a single object. (b) {P}romptable {C}oncept {S}egmentation (\textbf{PCS}) in SAM3 supports short NPs as prompts (\eg, ``\textit{soccer player}'') to segment all instances of a given concept. (c) To handle complex natural-language instructions, SAM3 relies on external multi-modal agents to convert instructions into short NPs, which remain overly coarse to capture fine-grained conditions ({\color{red}red icon}). (d) Promptable Instruction Segmentation (\textbf{PIS}) in SAM3-I directly interprets complex instructions and grounds the intended target instance ({\color{green}green icon}). SAM3-I introduces a unified framework, as the first of its kind in the SAM family, that integrates concept-level understanding and instruction-level reasoning.
    }
    \vspace{-.42cm}
    \label{fig:intro}
\end{figure*}

\begin{abstract}

Segment Anything Model 3 (SAM3) advances open-vocabulary segmentation through promptable concept segmentation, enabling users to segment all instances associated with a given concept using short noun-phrase (NP) prompts. While effective for concept-level grounding, real-world interactions often involve far richer natural-language instructions that combine attributes, relations, actions, states, or implicit reasoning. Currently, SAM3 relies on external multi-modal agents to convert complex instructions into NPs and conducts iterative mask filtering, leading to coarse representations and limited instance specificity. In this work, we present \textbf{\textit{SAM3-I}}, an instruction-following extension of the SAM family that unifies concept-level grounding and instruction-level reasoning within a single segmentation framework. Built upon SAM3, SAM3-I introduces an instruction-aware cascaded adaptation mechanism with dedicated alignment losses that progressively aligns expressive instruction semantics with SAM3's vision-language representations, enabling direct interpretation of natural-language instructions while preserving its strong concept recall ability. To enable instruction-following learning, we introduce \textbf{\textit{HMPL-Instruct}}, a large-scale instruction-centric dataset that systematically covers hierarchical instruction semantics and diverse target granularities. Experiments demonstrate that SAM3-I achieves appealing performance across referring and reasoning-based segmentation, showing that SAM3 can be effectively extended to follow complex natural-language instructions without sacrificing its original concept-driven strengths. Code and dataset are available at \url{https://github.com/debby-0527/SAM3-I}. 

\end{abstract}

\section{Introduction}
\vspace{-.09cm}
\label{sec:intro}

Semantic segmentation~\citep{sam,sam2,medsamadapter} is a fundamental task in visual understanding. Traditional methods~\citep{FCN2015,ji2023semanticrt} have primarily targeted explicit instance segmentation, where each pixel is assigned to one of a fixed set of predefined categories. However, such closed vocabularies remain considerably narrower than the diverse range of instances present in real-world scenarios. To this end, SAM3~\citep{sam3} represents an important step forward by introducing promptable concept segmentation (\textbf{PCS}), a concept-driven paradigm that enables users to segment all related instances using short NP prompts. 
As shown in Fig.~\ref{fig:intro}~b, a concept prompt such as ``soccer player'' can be used to segment all associated instances in an image.

However, real-world interactions are typically not conveyed through short NPs alone; users often describe instances using far richer expressions that may incorporate attributes, spatial relations, functions, actions, states, or even implicit reasoning over instances. Such descriptions often specify a concept together with multiple conditions, \eg, ``\textit{the soccer player on the far right, wearing a blue jersey, is sliding on the grass to tackle the ball}'', as shown in Fig.~\ref{fig:intro}~c. Handling such inputs requires the model to parse structured semantics and accurately ground them in the scene. This capability is essential for many open-world applications, including home robotics~\citep{quartey2025verifiably}, autonomous driving~\citep{feng2020deep}, and augmented reality~\citep{wahid2024augmented}, where systems must follow complex instructions to guide segmentation and decision-making.

This leads to a natural question: \textit{\textbf{Can we retain SAM3's strong concept recall ability gained from large-scale training while enabling it to interpret more complex instructions and ground the corresponding instances?}} To approximate this capability, SAM3 currently relies on external multi-modal agents, \eg, Qwen3-VL~\citep{Qwen3VL}, to convert long instructions into short NPs and then perform multiple rounds of mask filtering process. Although workable, this agent-driven pipeline separates linguistic reasoning from visual segmentation, adds notable computational overhead, and limits the model's ability to directly align complex instructions with visual concepts. More importantly, the PCS formulation constrains the agent's final output to short NPs. These NP-level concepts are overly coarse and often insufficient for precisely grounding specific instances described by detailed instructions, as illustrated in Fig.~\ref{fig:intro}~c and d.

To address these limitations, we propose \textbf{SAM3-I}, an instruction-following extension of the SAM family that unifies concept-level grounding and instruction-level reasoning within a single segmentation framework. Built upon SAM3, SAM3-I preserves its strong concept recall ability while enabling the direct interpretation of free-form natural-language instructions, ranging from simple referring expressions to complex reasoning-based queries. At its core, we introduce a dedicated instruction-aware cascaded adaptation mechanism with dedicated alignment losses that progressively aligns expressive instruction semantics with SAM3's existing vision-language representations. This design injects instruction-following capability in a parameter-efficient manner, without disrupting SAM3's original concept-driven strengths.

To support instruction-driven learning, we further construct \textbf{HMPL-Instruct}, a large-scale instruction-centric dataset derived from existing open-vocabulary segmentation benchmarks. HMPL-Instruct organizes natural-language instructions in a hierarchical semantic space, spanning concept-level prompts, explicit referring expressions, and complex reasoning-based instructions, while supporting diverse target granularities at both object and part levels. Moreover, the dataset also covers a wide range of instruction-target cardinalities, including \textit{one-to-one}, \textit{one-to-all}, and \textit{one-to-many} grounding scenarios, reflecting the diversity of real-world instructions. HMPL-Instruct is built via a human-in-the-loop pipeline that combines automatic instruction generation, agentic quality inspection, and targeted human correction, ensuring both scalability and precise visual grounding. 

Empirical results show that SAM3-I achieves appealing performance on both referring and reasoning segmentation, bringing robust instruction-following capability to the SAM family. We have released SAM3-I together with the HMPL-Instruct dataset and our in-house annotation pipeline to facilitate future research on instruction-based segmentation. Finally, we highlight several promising directions for further exploration.

\begin{figure*}
	\centering
    \vspace{-.5cm}
	\includegraphics[width=1\linewidth]{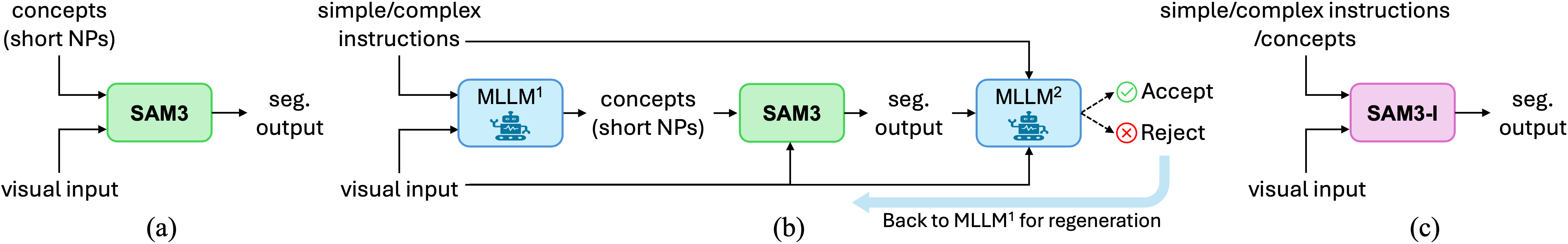}
    \vspace{-7mm}
	\caption{\textbf{Comparison of SAM3-series pipelines}. (a) SAM3 takes an image and a short NP as input and directly produces segmentation outputs. (b) SAM3 Agent handles longer instructions by prompting an MLLM$^1$ to interpret them into short NPs and using an MLLM$^2$ to validate and refine SAM3's predictions, often through multiple rounds of mask filtering. (MLLM$^1$ and MLLM$^2$ denote two separate calls to the same MLLM.) Although this enables SAM3 to process long instructions, it introduces additional system complexity, computational overhead, and may collapse detailed instructions into over-simplified NPs. (c) SAM3-I removes the agent loop entirely and directly supports both concepts and instructions in a single pass. For clarity, visual prompts (mask/point/box) are omitted.}
    \vspace{-.37cm}
	\label{fig:agent}
\end{figure*}

\vspace{-0.15cm}
\section{Preliminary: SAM Family}
\vspace{-0.15cm}
\label{sec:relatedwork}

SAM represents a major milestone in segmentation foundation models, demonstrating strong performance in diverse downstream applications, such as medical imaging and industrial inspection~\citep{medsamadapter,peng2025sam}. Its success is largely driven by the promptable segmentation paradigm introduced in SAM 1 \& 2~\citep{sam,sam2}, where users provide visual prompts (points/boxes/masks) to segment a single target per prompt, a setting known as Promptable Visual Segmentation (PVS) (Fig.~\ref{fig:intro}~a). Combined with large-scale, high-quality SA-1B and SA-V datasets, this paradigm forms the foundation of the SAM family's remarkable segmentation capability.

SAM3 extends this paradigm to PCS, enabling segmentation of all instances of a given concept using short NPs as prompts (e.g., ``\textit{soccer player}''), as shown in Fig.~\ref{fig:agent}~a. While this advances open-vocabulary segmentation, SAM3 is fundamentally designed around atomic concept prompts, restricting text inputs to simple NPs rather than multi-condition referring expressions or reasoning instructions. When presented with complex instructions, SAM3 relies on an agentic pipeline (Fig.~\ref{fig:agent}~b), where an external multi-modal large language model (MLLM) converts rich instructions into short NPs followed by iterative mask filtering. This multi-step process depends heavily on external reasoning and often oversimplifies natural-language descriptions, limiting precise instance-level grounding.

Real-world usage, however, involves rich, compositional instructions beyond simple NPs. This motivates us to extend SAM3 with the ability to follow complex instructions while preserving its strong concept-level grounding capability.

\begin{figure*}
	\centering
    \vspace{-.5cm}
	\includegraphics[width=.99\linewidth]{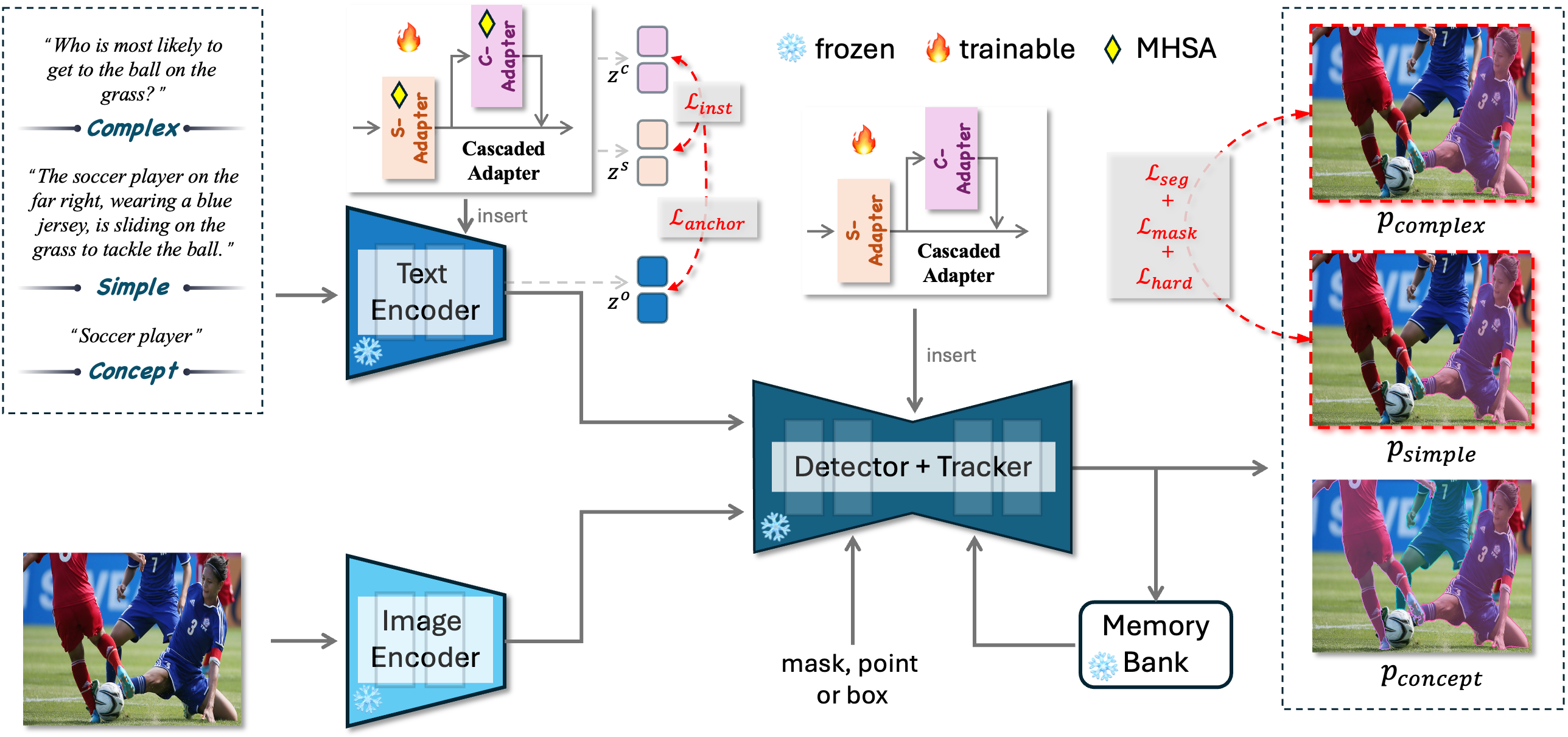}
    \vspace{-.15cm}
	\caption{\textbf{Overview of the proposed SAM3-I framework}. (Sec.~\ref{sec:model})}
	\label{fig:method}
    \vspace{-.43cm}
\end{figure*}

\vspace{-.1cm}
\section{Promptable Instruction Segmentation}
\vspace{-.1cm}
\label{sec:PIS}

We extend the PCS paradigm to a more expressive instruction-following setting, which we term Promptable Instruction Segmentation (\textbf{PIS}). We formulate instructions into three progressively finer hierarchical levels:
\begin{itemize}[leftmargin=*, topsep=1pt, itemsep=1pt]
    \item \textit{i) Concept instruction} represents the most atomic form, expressed as short NPs compatible with SAM3.
    \vspace{-.12cm}
    \item \textit{ii) Simple instruction} retains explicit NP mentions while incorporating additional referring conditions such as attributes, spatial relations, or local context.
    \vspace{-.12cm}
    \item \textit{iii) Complex instruction} removes explicit NP references and instead requires the model to identify targets by interpreting functions, actions, affordance, or broader contextual semantics.
\end{itemize}
These three levels establish a hierarchy that enables SAM-I to unify concept-level understanding and instruction-level reasoning within the SAM family.

\section{SAM3-I Model}
\vspace{-.1cm}
\label{sec:model}

\noindent\textbf{Overview.} As shown in Fig.~\ref{fig:method}, SAM3-I builds upon the SAM3 architecture and extends it to the PIS task, while remaining fully compatible with PCS and PVS. Briefly, SAM3 comprises a dual vision-language encoder and a detector-based segmentation head, along with a tracker and a memory module for video segmentation.

To equip SAM3 with the ability to interpret instructions from simple referring expressions to complex reasoning, we introduce an instruction-aware cascaded adaptation mechanism with distribution alignment losses, while keeping the SAM3 backbone frozen. This design preserves SAM3's original capabilities, prevents catastrophic forgetting, and injects new instruction-following ability in a parameter-efficient manner.

\noindent\textbf{Instruction-Aware Cascaded Adapter.} Instruction understanding spans a wide range of linguistic complexity. Simple referring expressions rely on explicit NPs combined with additional conditions, which requires the model to detect explicit linguistic anchors and align them with visual evidence. Complex instructions demand richer linguistic competence, requiring the model to interpret functional descriptions, implicit references, or perform contextual reasoning. Mastering this spectrum is essential for robust instruction-following segmentation.

To address this challenge, we introduce an instruction-aware cascaded adapter, a hierarchical module inserted into each layer of SAM3's text encoder. The design contains two adapters:
\begin{itemize}[leftmargin=*, topsep=2pt, itemsep=2pt]
    \item \textit{i) S-Adapter} learns attribute, position, and relation semantics, enabling effective grounding for simple referring expression where the target instance's NP remains explicitly available.
    \vspace{-.12cm}
    \item \textit{ii) C-Adapter} builds upon S-Adapter and is responsible for handling complex instructions that omit explicit target NP mentions and require contextual reasoning.
\end{itemize}

Both adapters employ a bottleneck structure (down-projection, {GELU} activation function, and up-projection) to enhance alignment with visual features, and a multi-head self-attention (MHSA) layer is incorporated to capture long-range textural dependencies. During training, S-Adapter is used exclusively for simple instructions, whereas complex instructions activate the full cascade so that C-Adapter can refine and enrich the representation produced by S-Adapter. This progressive design mirrors the natural hierarchy of linguistic difficulty, yielding structured linguistic embeddings that generalize across instruction forms. The cascaded adapter {(without MHSA)} is also integrated into SAM3's detector to propagate instruction semantics into the segmentation stage.

\noindent\textbf{Distribution Alignment Losses.} Allowing the two branches to interpret instructions independently may lead to \textit{semantic drift} between them. To maintain coherent grounding across different instruction levels, we introduce distribution alignment losses that regularize the two branches at both text embedding-level and output-level, beyond the segmentation loss $\mathcal{L}_{\text{seg}}$ used in SAM3. The alignment losses comprise four components as detailed below.

\textit{Instruction Contrastive Loss.} In SAM3-I, simple and complex instructions referring to the same target instance should yield consistent semantic representations, whereas instructions referring to different instances—even within the same concept category—should remain distinguishable. To achieve this, we introduce an instruction contrastive loss to regularize the adapter-enhanced text embeddings.

Specifically, for each training sample $i$, we treat the simple and complex text embeddings ($z_i^s$, $z_i^c$) extracted by the text encoder as a positive pair, and treat all instructions from other samples within a mini-batch as negatives. The instruction contrastive loss is defined as:
\setlength{\abovedisplayskip}{4pt}
\setlength{\belowdisplayskip}{4pt}
\setlength{\abovedisplayshortskip}{4pt}
\setlength{\belowdisplayshortskip}{4pt}
\begin{equation}
    \mathcal{L}_{\text{inst}}
    =
    \sum_i
    \left(
    \ell_{i}^{s \rightarrow c}
    +
    \ell_{i}^{c \rightarrow s}
    \right),
\end{equation}
\begin{align}
    \ell_i^{s\to c}
    &= -\log
    \frac{\exp(\langle z_i^s, z_i^c\rangle/\tau)}
    {\sum_k w_{ik}\exp(\langle z_i^s, z_k^c\rangle/\tau)},
    \label{equ:loss_stoc}
\end{align}
\vspace{-4pt}
where the weighting factor $w_{ik}$ is defined as
\begin{equation}
w_{ik} =
\begin{cases}
1, & y_k \neq y_i, \\
\rho, & y_k = y_i,
\end{cases}
\label{eq:ru}
\end{equation}
with $\rho < 1$ down-weighting negative samples from the same concept category to stabilize training, {and $y$ denotes the concept category.} {$\langle \cdot, \cdot\rangle$ denotes cosine similarity.} For clarity, we present only the formulation of $\ell_i^{s\to c}$ in Eq.~(\ref{equ:loss_stoc}), while $\ell_i^{c\to s}$ is defined symmetrically and omitted for brevity. This design encourages invariance across different instruction expressions of the same instance while retaining fine-grained discriminability among different instances.

\textit{Parent-rank Concept Anchoring Loss.} While the instance-level contrastive loss effectively structures the instruction embedding space, it does not explicitly constrain instructions to remain within their semantic concept boundaries. Directly aligning instruction embeddings with concept embeddings as in $\mathcal{L}_{\text{inst}}$ is undesirable, as concept prompts often describe all instances of a category, whereas instructions may refer to subsets.

To this end, we introduce a parent-rank concept anchoring loss that enforces a relative semantic constraint: an instruction embedding should be more similar to its own concept than to concepts from other categories. Specifically, for each instruction embedding $z_i^{sc}$ (averaged from the simple and complex text embeddings), we impose a margin-based ranking loss against concept embeddings from other categories:
\vspace{-.3cm}

{\small
\begin{equation}
{\mathcal{L}_{\text{anchor}}
=
\sum_i
\max \left(
0,\;
m
+
\max_{j:\, y_j \neq y_i} \left\langle z_i^{sc}, z_j^{o} \right\rangle
-
\left\langle z_i^{sc}, z_i^{o} \right\rangle
\right)},
\label{eq:anchor}
\end{equation}
} where $z_i^{o}$ is the corresponding concept embedding and $m$ is a margin hyper-parameter. Gradients are not propagated through the concept embeddings. This loss acts as a semantic guardrail, anchoring instruction embeddings to SAM3's original concept-level representation while preserving the flexibility required for subset-level instruction understanding.

\textit{Mask Distribution Alignment Loss.} To ensure that the simple- and complex-instruction branches produce compatible predictions for the same target, we align their mask distributions using a Kullback-Leibler (KL) divergence loss:
\begin{equation}
    {\mathcal{L}_{\text{mask}} =(\mathrm{KL}(p_{\text{s}} \;\Vert\; p_{\text{c}})+\mathrm{KL}(p_{\text{c}} \;\Vert\; p_{\text{s}}))/2}
\end{equation}
where $p_{\text{s}}$ and $p_{\text{c}}$ represent instruction-conditioned mask predictions for both simple and complex branches. This further encourages the two branches to share a consistent semantic space.

\textit{Uncertainty-Aware Hard-Region Supervision Loss.} Discrepancies between the two branches often arise in semantically difficult regions, \eg, occlusions, relational cues, or context-dependent areas. To emphasize learning on these challenging pixels, we compute an uncertainty map $w_{\text{uncertainty}}$ based on the disagreement between the simple- and complex-instruction predictions, measured using the Jensen-Shannon Divergence~\citep{lin1991divergence}. This uncertainty map is then used as an adaptive weight in an auxiliary cross-entropy ($\mathrm{CE}$) loss {between the mask predictions and ground-truth (GT)}:
\begin{equation}
    {\mathcal{L}_{\text{hard}} = w_{\text{uncertainty}}\cdot(\mathrm{CE}(p_{\text{s}}, GT)+\mathrm{CE}(p_{\text{c}}, GT)).}
\end{equation}
By guiding the model to focus on ambiguous or reasoning-intensive regions, this loss enhances robustness to complex linguistic instructions.

The overall training objective is $ \mathcal{L}_{\text{train}} =  \mathcal{L}_{\text{seg}} +  \mathcal{L}_{\text{inst}} + \mathcal{L}_{\text{anchor}} + \mathcal{L}_{\text{mask}} + \mathcal{L}_{\text{hard}}$.

\noindent\textbf{Multi-Stage Training.} SAM3-I is trained through a three-stage curriculum designed to ensure stable convergence and progressively acquire increasingly complex instruction-following capabilities.

In stage 1, the SAM3 backbone is frozen and the S-Adapter is trained on simple instructional data to establish category-, attribute-, and relation-level grounding ability. In stage 2, the S-Adapter is frozen, and the C-Adapter is initialized from it and trained with complex instructions to enable the model with functional, and reasoning-level grounding ability. Finally, in stage 3, all adapters are jointly activated and fine-tuned using the proposed alignment objectives, harmonizing the two branches and ensuring consistent instruction-conditioned predictions.

\begin{table*}[tbp]
  \centering
  \vspace{-.5cm}
  \caption{\textbf{Statistics of representative instruction-guided segmentation datasets.} The symbol $^\dagger$ indicates datasets that explicitly support one-to-many scenarios, as highlighted in the corresponding papers.}
  \vspace{-.2cm}
    \resizebox{!}{1.99 cm}{\begin{tabular}{lccccccccccc}
    \toprule
    \multirow{2}[2]{*}{\textbf{Dataset}} & \multicolumn{3}{c}{\textbf{Instruction Level}} & \multicolumn{3}{c}{\textbf{Instruction Granularity}} & \multirow{2}[2]{*}{\textbf{Obj.}} & \multirow{2}[2]{*}{\textbf{Part}} & \multirow{2}[2]{*}{\textbf{\#Samples}} & \multirow{2}[2]{*}{\textbf{\#Masks}} & \multirow{2}[2]{*}{\textbf{\#Instruct.}} \\
          & \textit{Concept} & \textit{Simple} & \textit{Complex} & \textit{1-to-1} & \textit{1-to-many}$^\dagger$ & \textit{1-to-all} &       &       &       &       &  \\
    \midrule
    RefCOCO~\citep{Kazemzadeh} & \cmark     & \cmark     &  \xxmark  &  \cmark      & \xxmark       &  \xxmark      & \cmark      & \xxmark       & 19,994      & 50,000      & 142,209 \\
    RefCOCO+~\citep{Kazemzadeh} & \cmark     & \cmark     & \xxmark     & \cmark      &  \xxmark     & \xxmark       & \cmark      & \xxmark      & 19,992 & 49,856 & 141,564 \\
    RefCOCOg~\citep{refcocog} & \cmark     & \cmark     & \xxmark     &  \cmark     &  \xxmark     &  \xxmark      & \cmark       & \xxmark      & 25,799      & 49,822      &95,010  \\
    gRefCOCO~\citep{grefcoco} & \cmark     & \cmark     & \xxmark    &  \cmark     & \cmark     &  \cmark      & \cmark       & \xxmark      & 19,994      & 60,287      & 278,232  \\
    Ref-ZOM~\citep{hu2023beyond} & \cmark     & \cmark     & \xxmark  & \cmark      & \cmark     &  \xxmark      & \cmark      & \xxmark       & {55,078}      & 74,942      & 90,199 \\
    ReasonSeg~\citep{lai2024lisa} & \xxmark     & \xxmark   & \cmark     &  \cmark      &  \xxmark     & \cmark      & \cmark       & \cmark      & 1,218      & 2,058      & 5,078 \\
    MMR~\citep{jang2025mmr}   & \cmark     & \xxmark   & \cmark     & \cmark      & \xxmark    &  \cmark     & \cmark     & \cmark     & 57,431      &  341,182     & 194,398 \\
    InstructPart~\citep{wan2025instructpart} & \cmark     & \cmark     & \cmark     & \cmark    & \xxmark    & \xxmark     & \xxmark    &  \cmark       & 2,400      & 2,400 & 14,400  \\
    \midrule
    \textbf{HMPL-Instruct} & \cmark     & \cmark     & \cmark     & \cmark     & \cmark     & \cmark     & \cmark    & \cmark     & 44,109 &  {347,120}   &  {849,792} \\
    \bottomrule
    \end{tabular}}
    \vspace{-.5cm}
  \label{tab:dataset}%
\end{table*}%

\section{HMPL-Instruct Dataset}
\label{sec:data}

In this section, we elaborate on the proposed \textit{Hierarchical Multi-grained PACO-LVIS-Instruct (HMPL-Instruct)} dataset.

\noindent\textbf{Motivation.} As summarized in Tab.~\ref{tab:dataset}, instruction-guided segmentation benchmarks have progressively expanded from simple referring expressions to more implicit, reasoning-oriented instructions. Along this evolution, three key attributes have emerged as increasingly important for practical instruction-following segmentation. First, real-world instructions often refer to multiple targets within a scene, requiring models to identify a specific subset or all instances of visually similar objects, rather than a single isolated instance. Second, instruction semantics naturally exhibit a hierarchical structure, spanning concept-level NPs, explicit referring expressions, and more complex instructions that rely on relational or functional reasoning without explicitly naming the target. Third, effective instruction following frequently requires perception at different semantic granularities, ranging from object-level understanding to fine-grained part-level recognition. These observations motivate the design of \textit{HMPL-Instruct} as a unified benchmark that jointly captures instruction hierarchy (concept, simple, and complex), multi-grained target cardinality (one-to-one, one-to-many, and one-to-all), and semantic granularity (object-level and part-level) within a single dataset.

\noindent\textbf{Dataset Construction.} To enable a unified formulation of hierarchical instructions and multi-grained targets, we develop a \textit{human-in-the-loop data engine} that enriches the PACO-LVIS dataset~\citep{paco} with compositional natural-language instructions. The pipeline has four stages (detailed diagram is provided in Appendix~\ref{Appx:dataset}):

\textit{Stage 1: Automatic Instruction Annotation.}
Given an image with instance masks and class labels, we construct a visual prompt comprising the original image, mask-overlaid image, and cropped masked regions. Conditioned on a system prompt and few-shot examples, an annotator MLLM generates diverse instruction candidates for each target instance, including simple instructions that explicitly mention the target NP and describe perceptual attributes, and complex instructions that omit the NP and rely on higher-level reasoning cues such as actions or functions. Both instruction types are produced in declarative and question formats. The annotator additionally synthesizes negative instructions by contradicting visual semantics, as well as a concise concept-level NP compatible with SAM3's PCS-style prompting. In this way, each target instance is associated with four positive instructions, four negative instructions, and one concept NP.

\textit{Stage 2: Agentic Quality Inspection.}
All generated instructions are verified by a second MLLM via a multiple-choice consistency check with the visual content and target mask. An instruction set is accepted only if the inspector achieves 100\% accuracy across all candidates; otherwise, it is rejected and returned to Stage~1 for regeneration, iteratively filtering semantic inconsistencies.

\begin{table*}[t]
\vspace{-.65cm}
\centering
\small
\caption{\textbf{Comparison of SAM3-I with SAM3 and SAM3 Agent baselines.} ``*'' indicates that we retain naive SAM3’s concept-level scores, and ``-'' denotes settings not applicable to the corresponding method.} 
\vspace{-.19cm}
\begin{tabular}{l c m{1.0cm}<{\centering}  m{1cm}<{\centering}m{1cm}<{\centering}  m{1cm}<{\centering}m{1cm}<{\centering}  m{1cm}<{\centering}m{1cm}<{\centering}} 
    & & & \multicolumn{2}{c}{\textit{Concept-level}}  & \multicolumn{2}{c}{\textit{Simple Instruct.}}  & \multicolumn{2}{c}{\textit{Complex Instruct.}} \\
    \cmidrule(lr){4-5}\cmidrule(lr){6-7}\cmidrule(lr){8-9}{Model} & {Parameter} & {Inference} & {\tt\small gIoU} & {\tt\small P@50} & {\tt\small gIoU} & {\tt\small P@50} & {\tt\small gIoU} & {\tt\small P@50} \\
    \toprule
    SAM3~\citep{sam3}          & 0.8B & single & 29.5 & 31.9 & -- & -- & -- & -- \\
    SAM3 Agent (Qwen3-VL-8B) & 8.8B & multi  & -- & -- & 28.4  & 28.9  & 26.4 &  27.4  \\
    \textbf{SAM3-I}        & 1.1B & single & 29.5$^{*}$ & 31.9$^{*}$ & {59.7} & {65.3} & {49.0} & {51.0} \\ 
    & & &  & 
    & \footnotesize{\textcolor{green!60!black}{+31.3}} & \footnotesize{\textcolor{green!60!black}{+36.4}}
    & \footnotesize{\textcolor{green!60!black}{+22.6}} & \footnotesize{\textcolor{green!60!black}{+23.6}} \\
\end{tabular}
\vspace{-.2cm}
\label{paco}
\end{table*}

\begin{figure*}
	\centering
    \vspace{-.1cm}
	\includegraphics[width=1\linewidth]{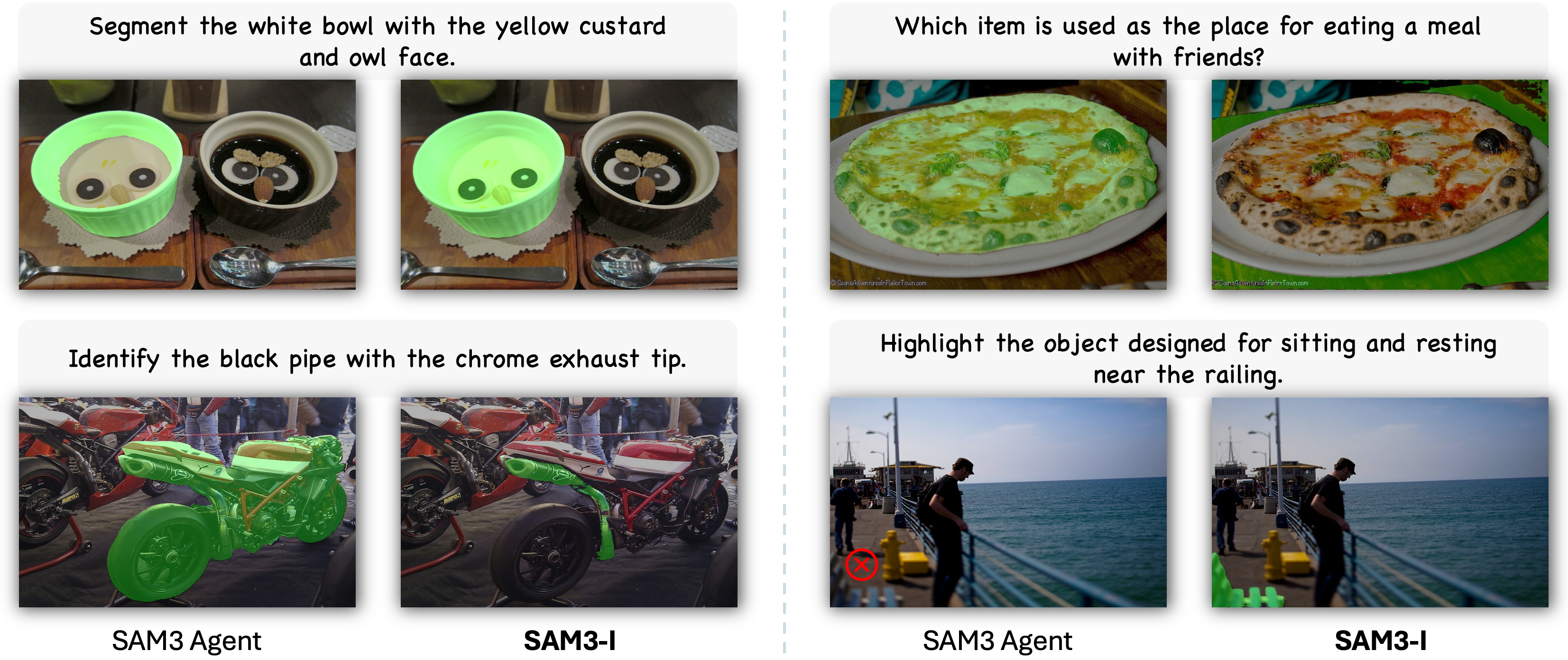}
    \vspace{-.77cm}
     \caption{\textbf{Qualitative comparison between SAM3-I and SAM3 Agent.} Here Qwen3-VL-8B is used for the agent.}
    \vspace{-.5cm}
	\label{fig:results}
\end{figure*}

\textit{Stage 3: Human-in-the-Loop Correction.}
Samples that repeatedly fail automatic verification due to ambiguous visuals or subtle linguistic cues are reviewed by three human annotators, who collaboratively revise or discard problematic instructions to ensure accurate and precise grounding. Through the first three stages, we obtain high-quality instruction-mask pairs that primarily support \textit{one-to-one} and \textit{one-to-all} grounding.

\textit{Stage 4: Human Annotation for One-to-Many Setting.}
The \textit{one-to-many} setting, which requires selecting a subset of visually similar instances, remains challenging for automatic MLLM-based annotation, as models often fail to consistently group subsets or generate unambiguous instructions. We therefore adopt a fully human-driven process using a dedicated web-based interface, allowing annotators to select instance subsets and associate them with corresponding simple and complex instructions. Ten annotators create initial annotations, and four inspectors refine them to ensure semantic correctness and compliance with predefined rules, such as excluding target NPs in complex instructions. Ambiguous cases are removed. Finally, the instructions are lightly polished using an MLLM-based rewriting step to improve fluency while preserving the original semantics. Using this process, we collect 2,941 one-to-many samples with 11,764 paired instructions and 7,449 masks, averaging 2.53 masks per image.

In summary, HMPL-Instruct comprises 133,960 object-level masks and 213,160 part-level masks across 44,109 images, paired with 849,792 positive instructions. The average instruction length is 15 words. We follow the original PACO-LVIS benchmark for the train/validation/test splits. Representative examples are provided in Appendix Fig.~\ref{fig:data1}.

\vspace{-0.2cm}
\section{Experiments}
\label{sec:exper}

\vspace{-0.2cm}
\subsection{Evaluation Metrics}

We evaluate our approach using {\tt\small gIoU} and {\tt\small P@50}. {\tt\small gIoU}~\citep{lai2024lisa} is computed as the average Intersection-over-Union (IoU) over all test images. {\tt\small P@50}~\citep{hu2016segmentation} reports the proportion of test samples whose predicted mask achieves an IoU of at least 0.5 with the ground-truth mask.

\vspace{-0.2cm}
\subsection{Quantitative Results}
\noindent\textbf{Comparison with SAM3 and SAM3 Agent.} We first assess the instruction-following ability of SAM3-I compared with the original SAM3 and SAM3 with MLLM agents on HMPL-Instruct. As shown in Tab.~\ref{paco}, SAM3 is limited to short NP concept prompts and only supports concept-level segmentation, while the agent-based variant enables instruction processing at the cost of multi-pass inference and substantially increased model size. In contrast, SAM3-I directly supports concept, simple, and complex instructions in a single-pass pipeline. Importantly, SAM3-I preserves SAM3's concept-level performance (29.5 {\tt\small gIoU} / 31.9 {\tt\small P@50}) {since we explicitly keep the SAM3 backbone frozen}, while achieving 59.7 / 65.3 on simple instructions and 49.0 / 51.0 on complex instructions. {This demonstrates that instruction-level learning does not degrade standard concept prompting. Moreover,} compared with the agent-based variant, SAM3-I attains appealing accuracy using fewer parameters and without relying on external MLLM reasoning. These results demonstrate that SAM3-I unifies concept grounding and instruction-level reasoning in an efficient extension of the SAM family.

\begin{table*}
  \vspace{-.5cm}
  \centering
  \small
  \caption{\textbf{Quantitative results on the HMPL-Instruct dataset under different target granularities}, including \textit{one-to-one}, \textit{one-to-many}, and \textit{one-to-all}, evaluated with simple and complex instructions.}
  \vspace{-.18cm}
  \resizebox{!}{1.55 cm}{\begin{tabular}{lcccccccccccc}
          & \multicolumn{6}{c}{\textit{Simple Instruct.}}          & \multicolumn{6}{c}{\textit{Complex Instruct.}} \\ \cmidrule(lr){2-7} \cmidrule(lr){8-13}
          & \multicolumn{2}{c}{\textit{1-to-1}} & \multicolumn{2}{c}{\textit{1-to-many}} & \multicolumn{2}{c}{\textit{1-to-all}} & \multicolumn{2}{c}{\textit{1-to-1}} & \multicolumn{2}{c}{\textit{1-to-many}} & \multicolumn{2}{c}{\textit{1-to-all}} \\  \cmidrule(lr){2-3} \cmidrule(lr){4-5} \cmidrule(lr){6-7} \cmidrule(lr){8-9} \cmidrule(lr){10-11} \cmidrule(lr){12-13}
    
    Model & {\tt\small gIoU} & {\tt\small P@50}   & {\tt\small gIoU} & {\tt\small P@50}   & {\tt\small gIoU} & {\tt\small P@50}   & {\tt\small gIoU} & {\tt\small P@50}   & {\tt\small gIoU} & {\tt\small P@50}   & {\tt\small gIoU} & {\tt\small P@50} \\ \toprule
    LISA~\citep{lai2024lisa}  &   24.1    &   20.2    &   21.8    &   14.4    &    19.4   & 11.8  &   18.2    &   13.8    &   18.2    &    10.8   &   17.8    & 10.2 \\
    UniPixel~\cite{liu2025unipixel} &   36.7    &  35.1     & 38.6  &    36.0   &   30.0    &   23.4    &    33.3   & 31.8  &    36.1   & 34.3  & 28.8  & 23.0 \\
    SAM3 Agent &  28.3   &  28.6 &   14.9    &    14.9   &   33.6    &   34.9    &  27.1  &   28.3    &  20.3 &   20.6    &   26.9    & 27.7 \\
    {\textbf{SAM3-I}}   & {60.8}  &   {66.7}    &  {48.9}     &    {50.0}  &   {60.9}   &   {67.8}     &   {48.1}    &   {49.6}    &    {39.9}   &   {39.5}    &     {54.7}  &  {58.7} \\
    \end{tabular}}
    \vspace{-.5cm}
  \label{tab:ourfine-grained}
\end{table*}%

\noindent\textbf{Analysis of Different Target Granularities.} 
Tab.~\ref{tab:ourfine-grained} evaluates instruction-following segmentation on HMPL-Instruct across three target granularities: \textit{one-to-one}, \textit{one-to-many}, and \textit{one-to-all}, under both simple and complex instructions.
SAM3-I achieves consistently strong performance across settings.
Notably, SAM3-I significantly outperforms the SAM3 agent-based baseline in the challenging \textit{one-to-many} setting, for both simple and complex instructions, demonstrating its ability to perform effective subset-level reasoning and instance discrimination among visually similar objects, rather than overfitting to category-level segmentation. Additional results at both object and part levels are provided in Appendix~\ref{Appx:results}.

\begin{table}[t]
    \centering
    \small
    \caption{\textbf{Results on the RefCOCO and Ref-ZOM}.}
    \vspace{-.2cm}
    \resizebox{!}{1.5 cm}{\begin{tabular}{l m{0.55cm}<{\centering} m{0.55cm}<{\centering} m{0.55cm}<{\centering}m{0.55cm}<{\centering}} 
        & \multicolumn{2}{c}{\textit{RefCOCO}}  & \multicolumn{2}{c}{\textit{Ref-ZOM}} \\
        \cmidrule(lr){2-3}\cmidrule(lr){4-5}{Model} & {\tt\small gIoU} & {\tt\small P@50} & {\tt\small gIoU} & {\tt\small P@50} \\
        \toprule
        LISA~\citep{lai2024lisa}            & 57.3 & 58.8 & 51.8 & 52.4 \\
        ReLA~\citep{grefcoco}           & 73.8 & 83.9 & 66.7 & 74.1 \\
        UniPixel~\cite{liu2025unipixel}     & {78.1} & {88.4} & 67.0 & 72.4 \\
        SAM3 Agent                 & 71.1 & 81.9 & 65.0 & 73.8 \\
        {\textbf{SAM3-I}}               & 76.3 & 85.5 & {73.6} & {80.0} \\
    \end{tabular}}
    \vspace{-0.2cm}
    \label{tab:otherdata}
\end{table}

\noindent\textbf{Evaluation on Representative Benchmarks.} We evaluate SAM3-I on two representative segmentation benchmarks, RefCOCO~\citep{Kazemzadeh} and Ref-ZOM~\citep{hu2023beyond}, in Tab.~\ref{tab:otherdata}. SAM3-I achieves competitive or superior performance, despite being trained with substantially less data (Tab.~\ref{tab:dataset}) than recent large-scale {non-agent} method UniPixel~\cite{liu2025unipixel}. This demonstrates that SAM3-I performs well to diverse instruction-driven benchmarks and highlights the benefits of building upon SAM3, a concept-driven segmentation foundation pretrained on large-scale pixel-level corpora, which provides robust and transferable concept representations for instruction-level extension.

\vspace{-.1cm}
\subsection{Qualitative Results}
Fig.~\ref{fig:results} provides a qualitative comparison between SAM3-I and SAM3 Agent (Qwen3-VL-8B) on instruction-following segmentation. The left column shows simple referring instructions, while the right column presents more challenging complex instructions. Across both settings, SAM3-I more accurately localizes the intended target than the agent-based baseline, while operating in a single-pass pipeline without external MLLM-driven multi-step mask filtering. For simple referring instructions, SAM3-I demonstrates stronger instance-level discrimination. For example, when asked to segment the ``\emph{black pipe with the chrome exhaust tip}'', SAM3 Agent incorrectly segments the entire motorcycle, whereas SAM3-I precisely isolates the specified component. For complex reasoning instructions, the advantage of SAM3-I is more evident. In the final example, SAM3 Agent fails to produce a valid segmentation, while SAM3-I correctly identifies the instructed object by directly grounding the implicit functional semantics.

Additional qualitative results are provided in Appendix~\ref{Appx:results}, covering a wide range of instruction-following scenarios, including one-to-one, one-to-many, and one-to-all settings, as well as object-level and part-level targets under both referring and reasoning-based instructions.

\begin{table}[t]
\centering
\caption{\textbf{Ablation studies of the proposed SAM3-I.} $^*$ denotes using a single adapter. $\mathcal{L}_{align}$ represents using all alignment losses.}
\vspace{-0.2cm}
\label{tab:ablation}
\setlength{\tabcolsep}{6pt}
\resizebox{!}{2.8 cm}{
\begin{tabular}{llcccc}
& & \multicolumn{2}{c}{\textit{Simple Instruct.}} 
& \multicolumn{2}{c}{\textit{Complex Instruct.}} \\
\cmidrule(lr){3-4} \cmidrule(lr){5-6}
Stages & Enabled Losses & {\tt gIoU} & {\tt P@50} & {\tt gIoU} & {\tt P@50} \\
\toprule
\multicolumn{2}{l}{Original SAM3} & - & - & - & - \\
\midrule
1 & $\mathcal{L}_{seg}$ & 57.6 & 62.0 & 32.9 & 31.0 \\

1,2 & $\mathcal{L}_{seg}$  & 57.6 & 62.0 & 45.9 & 46.5 \\

2 & $\mathcal{L}_{seg}$ & 40.8 & 40.5 & 43.5 & 44.7 \\

\midrule
1,2,3 & $\mathcal{L}_{seg} + \mathcal{L}_{inst}$ 
  & 58.6 & 64.1 & 47.4 & 49.1 \\

1,2,3 & $\mathcal{L}_{seg} + \mathcal{L}_{anchor}$ 
  & 58.6 & 63.8 & 47.1 & 48.0 \\

1,2,3 & $\mathcal{L}_{seg} + \mathcal{L}_{mask}$ 
  & 58.0 & 63.1 & 46.4 & 47.3 \\

1,2,3 & $\mathcal{L}_{seg} + \mathcal{L}_{hard}$ 
  & 58.1 & 63.4 & 46.7 & 48.3 \\

\midrule

1,2,3$^*$ & $\mathcal{L}_{seg} + \mathcal{L}_{align}$ 
  & 57.7 & 62.3 & 46.7 & 48.4 \\

3 & $\mathcal{L}_{seg} + \mathcal{L}_{align}$ & 57.3 & 62.0 & 45.4 & 46.4\\

\midrule
\textbf{1,2,3} & \textbf{SAM3-I (Ours)} 
  & {59.7} & {65.3} 
  & {49.0} & {51.0} \\
\end{tabular}}
\vspace{-0.5cm}
\label{ablation}
\end{table}

\vspace{-.1cm}
\subsection{Ablation Studies}
We ablate key components of SAM3-I on HMPL-Instruct, with results summarized in Tab.~\ref{ablation}.

\noindent\textbf{Multi-Stage Training of Cascaded Adapters.}
We begin with the original SAM3 as the baseline, which is fundamentally designed to support concept-level prompting but not compatible to handle either simple or complex natural-language instructions.
Introducing Stage~1 training enables the S-Adapter to learn attribute-, position-, and relation-level semantics from simple referring instructions. As shown in Tab.~\ref{ablation}, this stage yields appealing performance on simple instructions (57.6 {\tt\small gIoU} / 62.0 {\tt\small P@50}), while complex instruction performance remains undesired, as expected.
Extending training to Stage~2 further activates the C-Adapter and exposes the model to complex, NP-free instructions. This addition enables effective reasoning-based grounding, leading to clear improvements on complex instructions (45.9 {\tt\small gIoU} / 46.5 {\tt\small P@50}), while maintaining stable performance on simple ones. This progression confirms that complex instruction understanding cannot be acquired implicitly from simple supervision alone, but instead requires explicit reasoning-oriented training. 
{When only C-Adapter is introduced and trained directly on complex instructions, the model can learn a certain degree of reasoning capability, but consistently under-performs the full cascaded design in which the C-Adapter is initialized from the S-Adapter for complex instruction processing. This result suggests that attribute- and relation-level grounding learned by the S-Adapter provides an essential semantic foundation for subsequent high-level reasoning.}

\noindent\textbf{Effect of Alignment Losses.}
Building upon stage 1 and 2 early training, we further examine the effect of introducing alignment losses in Stage~3.
Adding the instruction contrastive loss ($\mathcal{L}_{\text{inst}}$) improves both simple and complex instruction performance (\eg, complex: 45.9 → 47.4 {\tt\small gIoU}), demonstrating its role in enforcing semantic consistency across different instruction formulations while preserving instance-level discrimination.
Incorporating the parent-rank concept anchoring loss ($\mathcal{L}_{\text{anchor}}$) also brings noticeable performance improvement (\eg, complex: 45.9 → 47.1 {\tt\small gIoU}), highlighting the importance of constraining instruction embeddings within appropriate concept boundaries.
At the prediction level, introducing the mask distribution alignment loss ($\mathcal{L}_{\text{mask}}$) yields consistent gains across instruction types (\eg, simple: 62.0 → 63.1 {\tt\small P@50}), indicating that aligning outputs from different instruction branches helps maintain coherent grounding.
Finally, adding the uncertainty-aware hard-region supervision loss ($\mathcal{L}_{\text{hard}}$) leads to performance gains (\eg, complex: 46.5 → 48.3 {\tt\small P@50}), demonstrating the importance of emphasizing semantically difficult regions to improve robustness. With all losses enabled, SAM3-I achieves the best overall performance (59.7 {\tt\small gIoU} / 65.3 {\tt\small P@50}) on simple instructions and (49.0 / 51.0) on complex instructions, demonstrating that the alignment objectives play complementary roles in regulating instruction semantics and enforcing consistent and robust visual grounding.

\noindent\textbf{Cascaded Adapters \textit{vs.} Single Adapter.} We further validate the necessity of the cascaded design by comparing it with a single-adapter variant (marked as $^*$ in Tab. \ref{ablation}). It is observed that a single-adapter variant trained with the same curriculum and alignment losses under-performs the cascaded S-/C-Adapter design (\eg, complex: 46.7 \textit{vs.} 49.0 {\tt\small gIoU}). This suggests that explicitly separating linguistic roles enables progressive refinement of structured, layered instruction representations—an ability that a single adapter struggles to acquire within a shared parameter space.

\noindent\textbf{Multi-stage \textit{vs.} Single-stage.} We further remove the proposed staged curriculum and instead train the cascaded adapters with all losses in a single stage. This results in clear performance degradation on both simple and complex instructions. Without progressive training, the model is required to learn attribute grounding, relation modeling, and high-level reasoning simultaneously, which makes optimization substantially more difficult and prevents the cascaded adapters from forming well-structured, hierarchical linguistic representations.

Ablation results for hyper-parameters $\rho$ and $m$ in Eq.~(\ref{eq:ru}) and Eq.~(\ref{eq:anchor}) are provided in Appendix~\ref{Appx:results}.

\section{Conclusion}
\vspace{-.05cm}
In this work, we presented SAM3-I, an instruction-following extension of the SAM family that unifies concept-level grounding and instruction-level reasoning within a single segmentation framework. Supported by a scalable instruction-centric data construction pipeline, SAM3-I demonstrates that SAM3 can be effectively extended to follow natural-language instructions while preserving its strong concept-driven segmentation capability. We further introduce HMPL-Instruct, a dataset that systematically covers hierarchical instruction semantics and diverse target granularities, thereby providing a unified benchmark for instruction-following segmentation. We hope that SAM3-I serves as an initial yet meaningful step toward more capable and general instruction-driven segmentation foundation models, and encourages future exploration in this emerging direction.

\section*{Acknowledgment}
Thanks to the anonymous reviewers for their helpful feedback. This work is partly supported by NSERC Discovery, CFI-JELF, NSERC Alliance, Alberta Innovates, PrairiesCan grants, {Shenzhen Science and Technology Program under Grant No.~JCYJ20250604182948064} and Tencent Project UP. The views and conclusions contained in this paper are those of the authors and should not be interpreted as representing any funding agency.

\section*{Limitations}
\label{sec:discuss}

While SAM3-I demonstrates the feasibility of extending the SAM family toward instruction-following segmentation, several limitations and opportunities for further improvement remain.

\textit{i) Scaling training data for richer instruction knowledge.} Our data engine offers a prototype for generating rich instructional corpus, yet scaling to larger and more diverse instruction-mask datasets would further strengthen the model's ability to understand and reason over natural-language instructions. Similar to how large-scale datasets such as SA-1B, SA-V, and SA-Co~\citep{sam,sam2,sam3} fueled the evolution of the SAM family, scaling PIS data across broader object categories, part-level annotations, and reasoning patterns represents an important step toward a more comprehensive instruction-grounded segmentation foundation model.

\textit{ii) Potential bias of instruction-instance pairs.} Existing instruction-based segmentation benchmarks are still dominated by \textit{one-to-one} and \textit{one-to-all} instruction-instance mappings, where a single instruction refers to a single instance or all instances of a given concept. In contrast, more challenging \textit{one-to-many} settings, where an instruction simultaneously refers to multiple specific instances, have received comparatively less attention. This imbalance introduces a data bias that may limit the model's generalization ability to group-level or collective reasoning scenarios. Addressing this limitation calls for curating data with more diverse instruction-instance cardinalities and explicitly modeling multi-instance grounding in training.

\textit{iii) Beyond instruction-driven segmentation toward higher-level understanding.} Following the design philosophy of the SAM family, SAM3-I primarily focuses on enhancing segmentation foundations through instruction-driven prompting, without explicitly integrating high-level dense grounded understanding or conversational capabilities. While the current design choice allows SAM3-I to remain lightweight and modular, it also leaves room for future extensions. An interesting direction is to couple instruction-following segmentation with deeper semantic reasoning, or conversational feedback, potentially by integrating large language models with stronger contextual understanding (\eg, dense grounded understanding frameworks such as Sa2VA~\cite{sa2va}). Such extensions could enable more comprehensive human-model interaction while preserving SAM3-I's powerful segmentation foundation.

\section*{Ethical Considerations}

The proposed SAM3-I framework and the accompanying dataset are designed to support research on instruction-based segmentation and do not involve any content that violates ethical norms or societal standards. The proposed HMPL-Instruct dataset does not contain personally identifiable information and is not intended for surveillance, identification, or privacy-invasive applications. We envisage that the most proximate impacts of this work will be positive, providing an open and valuable resource for research community.

\bibliography{custom}

@inproceedings{Kazemzadeh,
  title={Referitgame: Referring to objects in photographs of natural scenes},
  author={Kazemzadeh, Sahar and Ordonez, Vicente and Matten, Mark and Berg, Tamara},
  booktitle={Proceedings of the Conference on Empirical Methods in Natural Language Processing (EMNLP)},
  pages={787--798},
  year={2014}
}

@inproceedings{chen2020simple,
  title={A simple framework for contrastive learning of visual representations},
  author={Chen, Ting and Kornblith, Simon and Norouzi, Mohammad and Hinton, Geoffrey},
  booktitle={International Conference on Machine Learning},
  pages={1597--1607},
  year={2020}
}

@inproceedings{liu2025unipixel,
  title={UniPixel: Unified Object Referring and Segmentation for Pixel-Level Visual Reasoning},
  author={Liu, Ye and Ma, Zongyang and Pu, Junfu and Qi, Zhongang and Wu, Yang and Ying, Shan and Chen, Chang Wen},
  booktitle={Advances in Neural Information Processing Systems},
  year={2025}
}

@article{sa2va,
  title={Sa2va: Marrying sam2 with llava for dense grounded understanding of images and videos},
  author={Yuan, Haobo and Li, Xiangtai and Zhang, Tao and Sun, Yueyi and Huang, Zilong and Xu, Shilin and Ji, Shunping and Tong, Yunhai and Qi, Lu and Feng, Jiashi and Yang, Ming-Hsuan},
  journal={arXiv preprint arXiv:2501.04001},
  year={2025}
}

@inproceedings{wan2025instructpart,
  title={Instructpart: Task-oriented part segmentation with instruction reasoning},
  author={Wan, Zifu and Xie, Yaqi and Zhang, Ce and Lin, Zhiqiu and Wang, Zihan and Stepputtis, Simon and Ramanan, Deva and Sycara, Katia P},
  booktitle={Proceedings of the Annual Meeting of the Association for Computational Linguistics},
  pages={24202--24227},
  year={2025}
}

@inproceedings{hu2023beyond,
  title={Beyond one-to-one: Rethinking the referring image segmentation},
  author={Hu, Yutao and Wang, Qixiong and Shao, Wenqi and Xie, Enze and Li, Zhenguo and Han, Jungong and Luo, Ping},
  booktitle={Proceedings of the IEEE/CVF International Conference on Computer Vision},
  pages={4067--4077},
  year={2023}
}

@inproceedings{grefcoco,
  title={GRES: Generalized referring expression segmentation},
  author={Liu, Chang and Ding, Henghui and Jiang, Xudong},
  booktitle={Proceedings of the IEEE/CVF Conference on Computer Vision and Pattern Recognition},
  pages={23592--23601},
  year={2023}
}

@inproceedings{refcocog,
  title={Modeling context in referring expressions},
  author={Yu, Licheng and Poirson, Patrick and Yang, Shan and Berg, Alexander C and Berg, Tamara L},
  booktitle={European Conference on Computer Vision},
  pages={69--85},
  year={2016}
}

@article{jang2025mmr,
  title={MMR: A large-scale benchmark dataset for multi-target and multi-granularity reasoning segmentation},
  author={Jang, Donggon and Cho, Yucheol and Lee, Suin and Kim, Taehyeon and Kim, Dae-Shik},
  journal={International Conference on Learning Representations},
  year={2025}
}

@inproceedings{quartey2025verifiably,
  title={Verifiably following complex robot instructions with foundation models},
  author={Quartey, Benedict and Rosen, Eric and Tellex, Stefanie and Konidaris, George},
  booktitle={IEEE International Conference on Robotics and Automation},
  pages={1--8},
  year={2025},
  organization={IEEE}
}

@inproceedings{wahid2024augmented,
  title={Augmented reality model in supporting instruction process: a critical review},
  author={Wahid, Azhar and Huda, Miftachul and Rohim, Moh Abdul and Ali, Abdul Halim and Kaspin, Khairul Ghufran and Fiqiyah, Maskanatul and Jima’ain, Muhammad Talhah Ajmain},
  booktitle={International Congress on Information and Communication Technology},
  pages={69--83},
  year={2024},
  organization={Springer}
}

@article{lin1991divergence,
  title={Divergence measures based on the Shannon entropy},
  author={Lin, Jianhua},
  journal={IEEE Transactions on Information Theory},
  volume={37},
  number={1},
  pages={145--151},
  year={1991},
  publisher={IEEE}
}

@inproceedings{FCN2015,
  title={Fully convolutional networks for semantic segmentation},
  author={Long, Jonathan and Shelhamer, Evan and Darrell, Trevor},
  booktitle={Proceedings of the IEEE/CVF Conference on Computer Vision and Pattern Recognition},
  pages={3431--3440},
  year={2015}
}

@inproceedings{hu2016segmentation,
  title={Segmentation from natural language expressions},
  author={Hu, Ronghang and Rohrbach, Marcus and Darrell, Trevor},
  booktitle={European Conference on Computer Vision},
  pages={108--124},
  year={2016}
}

@inproceedings{ji2023semanticrt,
  title={Semanticrt: A large-scale dataset and method for robust semantic segmentation in multispectral images},
  author={Ji, Wei and Li, Jingjing and Bian, Cheng and Zhang, Zhicheng and Cheng, Li},
  booktitle={Proceedings of the 31st ACM International Conference on Multimedia},
  pages={3307--3316},
  year={2023}
}

@inproceedings{lai2024lisa,
  title={Lisa: Reasoning segmentation via large language model},
  author={Lai, Xin and Tian, Zhuotao and Chen, Yukang and Li, Yanwei and Yuan, Yuhui and Liu, Shu and Jia, Jiaya},
  booktitle={Proceedings of the IEEE/CVF Conference on Computer Vision and Pattern Recognition},
  pages={9579--9589},
  year={2024}
}

@inproceedings{paco,
  title={Paco: Parts and attributes of common objects},
  author={Ramanathan, Vignesh and Kalia, Anmol and Petrovic, Vladan and Wen, Yi and Zheng, Baixue and Guo, Baishan and Wang, Rui and Marquez, Aaron and Kovvuri, Rama and Kadian, Abhishek and Mousavi, Amir and Song, Yiwen and Dubey, Abhimanyu and Mahajan, Dhruv},
  booktitle={Proceedings of the IEEE/CVF Conference on Computer Vision and Pattern Recognition},
  pages={7141--7151},
  year={2023}
}

@article{peng2025sam,
  title={Sam-lad: Segment anything model meets zero-shot logic anomaly detection},
  author={Peng, Yun and Lin, Xiao and Ma, Nachuan and Du, Jiayuan and Liu, Chuangwei and Liu, Chengju and Chen, Qijun},
  journal={Knowledge-Based Systems},
  volume={314},
  pages={113176},
  year={2025},
  publisher={Elsevier}
}

@article{medsamadapter,
title = {Medical SAM adapter: Adapting segment anything model for medical image segmentation},
journal = {Medical Image Analysis},
volume = {102},
pages = {103547},
year = {2025},
issn = {1361-8415},
author = {Junde Wu and Ziyue Wang and Mingxuan Hong and Wei Ji and Huazhu Fu and Yanwu Xu and Min Xu and Yueming Jin},
}

@article{sam2,
  title={Sam 2: Segment anything in images and videos},
  author={Ravi, Nikhila and Gabeur, Valentin and Hu, Yuan-Ting and Hu, Ronghang and Ryali, Chaitanya and Ma, Tengyu and Khedr, Haitham and R{\"a}dle, Roman and Rolland, Chloe and Gustafson, Laura and Mintun, Eric and Pan, Junting and Alwala, Kalyan Vasudev and Carion, Nicolas and Wu, Chao-Yuan and Girshick, Ross and Doll{\'a}r, Piotr and Feichtenhofer, Christoph},
  journal={International Conference on Learning Representations},
  year={2025}
}

@inproceedings{sam,
  title={Segment anything},
  author={Kirillov, Alexander and Mintun, Eric and Ravi, Nikhila and Mao, Hanzi and Rolland, Chloe and Gustafson, Laura and Xiao, Tete and Whitehead, Spencer and Berg, Alexander C. and Lo, Wan-Yen and Dollar, Piotr and Girshick, Ross},
  booktitle={Proceedings of the IEEE/CVF International Conference on Computer Vision},
  pages={4015--4026},
  year={2023}
}

@article{sam3,
      title={SAM 3: Segment Anything with Concepts}, 
      author={Nicolas Carion and Laura Gustafson and Yuan-Ting Hu and Shoubhik Debnath and Ronghang Hu and Didac Suris and Chaitanya Ryali and Kalyan Vasudev Alwala and Haitham Khedr and Andrew Huang and Jie Lei and Tengyu Ma and Baishan Guo and Arpit Kalla and Markus Marks and Joseph Greer and Meng Wang and Peize Sun and Roman Rädle and Triantafyllos Afouras and Effrosyni Mavroudi and Katherine Xu and Tsung-Han Wu and Yu Zhou and Liliane Momeni and Rishi Hazra and Shuangrui Ding and Sagar Vaze and Francois Porcher and Feng Li and Siyuan Li and Aishwarya Kamath and Ho Kei Cheng and Piotr Dollár and Nikhila Ravi and Kate Saenko and Pengchuan Zhang and Christoph Feichtenhofer},
      year={2025},
      journal={arXiv preprint arXiv:2511.16719},
}

@article{Qwen3VL,
  title={Qwen3-VL Technical Report}, 
  author={Shuai Bai and Yuxuan Cai and Ruizhe Chen and Keqin Chen and Xionghui Chen and Zesen Cheng and Lianghao Deng and Wei Ding and Chang Gao and Chunjiang Ge and Wenbin Ge and Zhifang Guo and Qidong Huang and Jie Huang and Fei Huang and Binyuan Hui and Shutong Jiang and Zhaohai Li and Mingsheng Li and Mei Li and Kaixin Li and Zicheng Lin and Junyang Lin and Xuejing Liu and Jiawei Liu and Chenglong Liu and Yang Liu and Dayiheng Liu and Shixuan Liu and Dunjie Lu and Ruilin Luo and Chenxu Lv and Rui Men and Lingchen Meng and Xuancheng Ren and Xingzhang Ren and Sibo Song and Yuchong Sun and Jun Tang and Jianhong Tu and Jianqiang Wan and Peng Wang and Pengfei Wang and Qiuyue Wang and Yuxuan Wang and Tianbao Xie and Yiheng Xu and Haiyang Xu and Jin Xu and Zhibo Yang and Mingkun Yang and Jianxin Yang and An Yang and Bowen Yu and Fei Zhang and Hang Zhang and Xi Zhang and Bo Zheng and Humen Zhong and Jingren Zhou and Fan Zhou and Jing Zhou and Yuanzhi Zhu and Ke Zhu},
  journal={arXiv preprint arXiv:2511.21631},
  year={2025}
}

@article{feng2020deep,
  title={Deep multi-modal object detection and semantic segmentation for autonomous driving:Datasets,methods,and challenges},
  author={Feng, Di and Haase-Sch{\"u}tz, Christian and Rosenbaum, Lars and Hertlein, Heinz and Glaeser, Claudius and Timm, Fabian and Wiesbeck, Werner and Dietmayer, Klaus},
  journal={IEEE Transactions on Intelligent Transportation Systems},
  volume={22},
  pages={1341--1360},
  year={2020},
  publisher={IEEE}
}

@inproceedings{radford2021learning,
  title={Learning transferable visual models from natural language supervision},
  author={Radford, Alec and Kim, Jong Wook and Hallacy, Chris and Ramesh, Aditya and Goh, Gabriel and Agarwal, Sandhini and Sastry, Girish and Askell, Amanda and Mishkin, Pamela and Clark, Jack and Krueger, Gretchen and Sutskever, Ilya},
  booktitle={International Conference on Machine Learning},
  pages={8748--8763},
  year={2021},
  organization={PmLR}
}

@inproceedings{liang2023open,
  title={Open-vocabulary semantic segmentation with mask-adapted clip},
  author={Liang, Feng and Wu, Bichen and Dai, Xiaoliang and Li, Kunpeng and Zhao, Yinan and Zhang, Hang and Zhang, Peizhao and Vajda, Peter and Marculescu, Diana},
  booktitle={Proceedings of the IEEE/CVF conference on computer vision and pattern recognition},
  pages={7061--7070},
  year={2023}
}

@inproceedings{sun2023going,
  title={Going denser with open-vocabulary part segmentation},
  author={Sun, Peize and Chen, Shoufa and Zhu, Chenchen and Xiao, Fanyi and Luo, Ping and Xie, Saining and Yan, Zhicheng},
  booktitle={Proceedings of the IEEE/CVF International Conference on Computer Vision},
  pages={15453--15465},
  year={2023}
}

@inproceedings{ouyang2023slvit,
  title={SLViT: Scale-Wise Language-Guided Vision Transformer for Referring Image Segmentation.},
  author={Ouyang, Shuyi and Wang, Hongyi and Xie, Shiao and Niu, Ziwei and Tong, Ruofeng and Chen, Yen-Wei and Lin, Lanfen},
  booktitle={International Joint Conference on Artificial Intelligence},
  pages={1294--1302},
  year={2023}
}

@inproceedings{yang2022lavt,
  title={Lavt: Language-aware vision transformer for referring image segmentation},
  author={Yang, Zhao and Wang, Jiaqi and Tang, Yansong and Chen, Kai and Zhao, Hengshuang and Torr, Philip HS},
  booktitle={Proceedings of the IEEE/CVF Conference on Computer Vision and Pattern Recognition},
  pages={18155--18165},
  year={2022}
}

@inproceedings{zou2023generalized,
  title={Generalized decoding for pixel, image, and language},
  author={Zou, Xueyan and Dou, Zi-Yi and Yang, Jianwei and Gan, Zhe and Li, Linjie and Li, Chunyuan and Dai, Xiyang and Behl, Harkirat and Wang, Jianfeng and Yuan, Lu and Peng, Nanyun and Wang, Lijuan and Lee, Yong Jae and Gao, Jianfeng},
  booktitle={Proceedings of the IEEE/CVF Conference on Computer Vision and Pattern Recognition},
  pages={15116--15127},
  year={2023}
}

@article{zou2023segment,
  title={Segment everything everywhere all at once},
  author={Zou, Xueyan and Yang, Jianwei and Zhang, Hao and Li, Feng and Li, Linjie and Wang, Jianfeng and Wang, Lijuan and Gao, Jianfeng and Lee, Yong Jae},
  journal={Advances in Neural Information Processing Systems},
  volume={36},
  pages={19769--19782},
  year={2023}
}

@inproceedings{xu2023side,
  title={Side adapter network for open-vocabulary semantic segmentation},
  author={Xu, Mengde and Zhang, Zheng and Wei, Fangyun and Hu, Han and Bai, Xiang},
  booktitle={Proceedings of the IEEE/CVF Conference on Computer Vision and Pattern Recognition},
  pages={2945--2954},
  year={2023}
}

@article{shin2024towards,
  title={Towards open-vocabulary semantic segmentation without semantic labels},
  author={Shin, Heeseong and Kim, Chaehyun and Hong, Sunghwan and Cho, Seokju and Arnab, Anurag and Seo, Paul Hongsuck and Kim, Seungryong},
  journal={Advances in Neural Information Processing Systems},
  volume={37},
  pages={9153--9177},
  year={2024}
}

\newpage
\appendix
\section*{Appendix}
\vspace{.1cm}
\label{sec:appendix}

This appendix provides supplementary material supporting the main paper. We first review recent efforts on language-guided segmentation and instruction-centric segmentation benchmarks related to SAM3-I in Appendix~\ref{Appx:relatedwork}. We then present additional details on dataset construction in Appendix~\ref{Appx:dataset}, followed by extended experimental results in Appendix~\ref{Appx:results}, including hyperparameter analysis, additional visualizations, and further qualitative and quantitative evaluations. Implementation details are provided in Appendix~\ref{Appx:imple}. Finally, Appendix~\ref{Appx:ethics} presents statements on artifacts and ethics.

\vspace{.15cm}
\section{Related Work}
\vspace{.15cm}
\label{Appx:relatedwork}

\subsection{Evolving Language-Guided Segmentation}
Segmenting objects described by language has attracted increasing attention in recent years, driven by the demand for flexible, open-ended, and human-centric visual understanding systems. Along this line, prior studies can be broadly categorized by the level of linguistic abstraction they support.

Early efforts primarily focus on \textit{open-vocabulary semantic segmentation}, which aims to generalize segmentation models beyond a fixed closed set of categories~\cite{liang2023open,sun2023going,xu2023side,shin2024towards} by leveraging large-scale vision-language pretraining~\cite{radford2021learning}. By aligning visual representations with textual embeddings learned from image-text corpora, these approaches enable open-set category recognition and segmentation, laying an important foundation for language-driven segmentation. However, most open-vocabulary methods are limited to short NPs or category names, and are not designed to handle richer instructional language or instance-specific reasoning.

To support more precise instance-level grounding, \textit{referring expression segmentation}~\cite{hu2016segmentation} is introduced to segment a target object specified by an explicit natural language description. A large body of work~\cite{ouyang2023slvit,yang2022lavt,grefcoco,hu2023beyond} explores cross-modal feature fusion, attention mechanisms, and relational modeling to resolve attributes, spatial relations, and inter-object dependencies. More recent approaches~\cite{zou2023generalized,zou2023segment,sa2va} further benefit from vision and vision-language foundation models, such as SAM and CLIP, significantly improving robustness and generalization. Despite these advances, referring expression segmentation typically assumes that the target object is explicitly mentioned and that each instruction corresponds to a single, well-defined instance, limiting its applicability to more implicit or compositional instructions.

More recently, research~\cite{lai2024lisa} has begun to explore \textit{reasoning-based segmentation}, where the target is specified implicitly through functional, relational, or commonsense descriptions, rather than explicit object names, marking an important step toward higher-level semantic understanding. Following efforts~\cite{wan2025instructpart,jang2025mmr} investigate part-level segmentation with instruction reasoning, enhancing the ability of segmentation models to recognize each part of the target and its corresponding affordances before grounding to task-related regions.

Overall, the evolution from category-level open-vocabulary segmentation, to instance-level referring expression segmentation, and further toward reasoning-centric segmentation reveals a clear trend toward richer language understanding and more flexible visual grounding. In parallel, the SAM family has evolved from relying on low-level visual prompts to support concept-level prompts, demonstrating strong generalization and robust concept understanding across diverse domains. Nevertheless, existing SAM-based models are not explicitly designed to follow free-form natural language instructions or perform instruction-level reasoning. These observations motivate us to extend the SAM paradigm toward \textit{instruction-following segmentation}, leading to \textbf{SAM3-I}, which aims to unify concept-level understanding and instruction-level reasoning within the SAM family.

\begin{figure*}[ht!]
	\centering
    \vspace{.3cm}
	\includegraphics[width=1\linewidth]{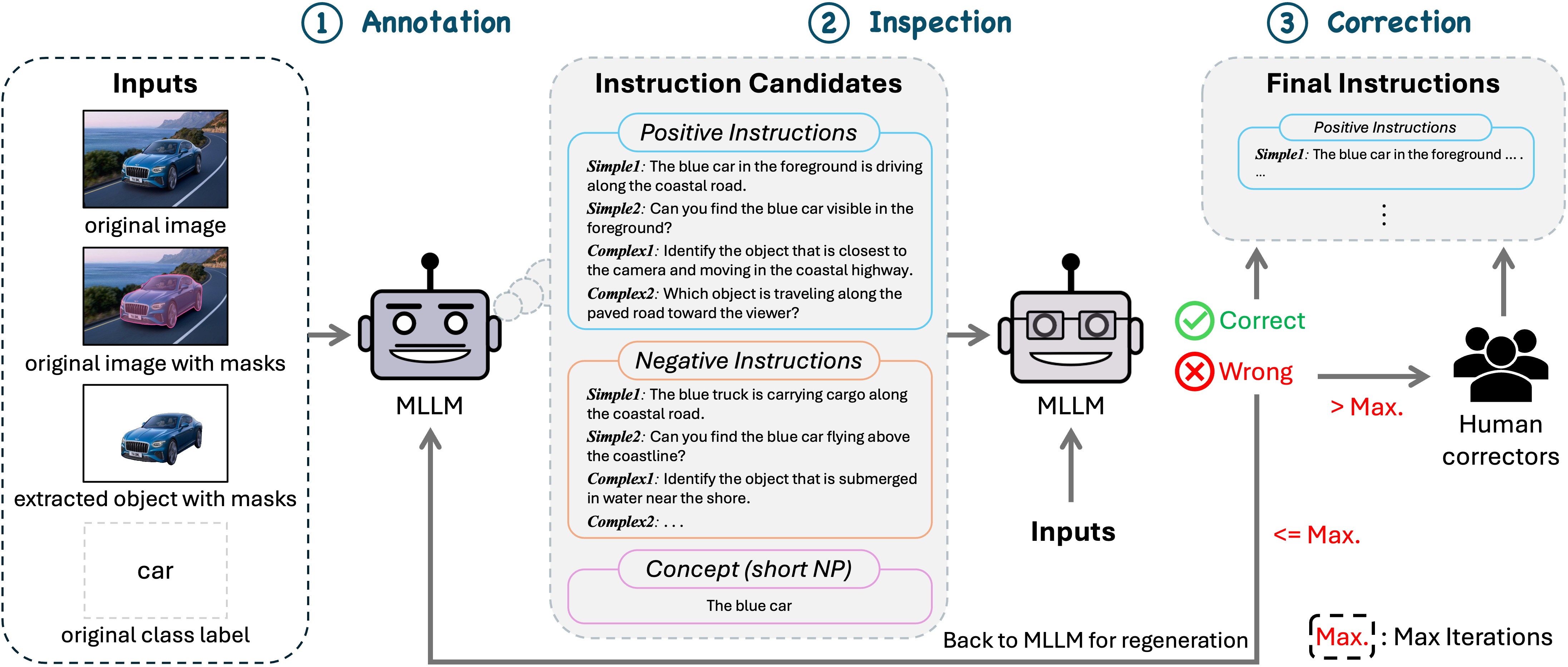}
	\caption{\textbf{Overview of the scalable instructional data construction pipeline}. (Sec.~\ref{sec:data})}
	\label{fig:engine}
\end{figure*}

\subsection{Instruction-Centric Segmentation Benchmarks}
A variety of datasets have been proposed to support referring expression and reasoning-based segmentation. Early benchmarks such as RefCOCO, RefCOCO+, and RefCOCOg~\cite{Kazemzadeh, refcocog} focus on explicit object references with increasing linguistic complexity, encouraging models to reason over attributes and relations rather than relying solely on location cues. Subsequent extensions~\cite{grefcoco,hu2023beyond} explore multi-target referring scenarios, but still assume that target categories are explicitly specified in the text.

More recent datasets shift toward implicit and reasoning-driven instructions, where target names are intentionally omitted and segmentation relies on functional or semantic descriptions~\cite{lai2024lisa}. This dataset significantly advances the field by introducing higher-level reasoning requirements, and have been further extended to part-level and multi-granularity settings~\cite{wan2025instructpart,jang2025mmr}. While these efforts substantially enrich instruction diversity, they typically address specific referring or reasoning paradigms and target scopes in isolation.

In contrast, our \textbf{HMPL-Instruct} dataset aims to provide a unified benchmark that systematically covers the full spectrum of instruction types. It integrates concept-level prompts, explicit referring expressions, and implicit complex instructions within a single dataset, while supporting object-level and part-level annotations as well as diverse target cardinalities, ranging from one-to-one and one-to-many to one-to-all scenarios. By unifying instruction hierarchy and target granularity in a coherent benchmark, HMPL-Instruct is designed to facilitate the development and evaluation of general instruction-following segmentation models.

\begin{figure*}
	\centering
    \vspace{.2cm}
	\includegraphics[width=1\linewidth]{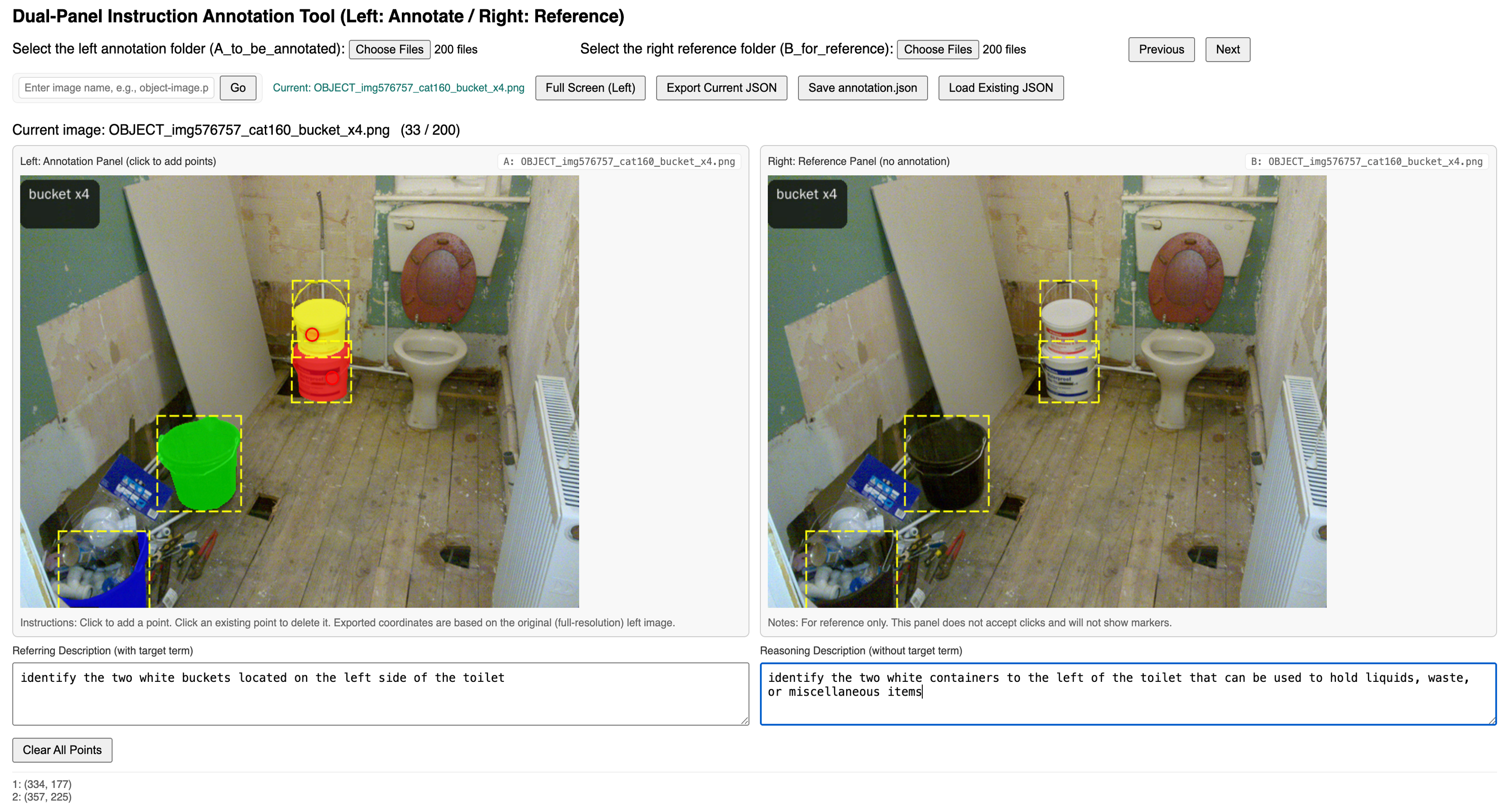}
    \caption{\textbf{Overview of our dual-panel instruction annotation tool.} The interface presents the same image in two synchronized panels. The left \emph{Annotate} panel supports interactive point placement with mask overlays to specify target instances or subsets (e.g., \textit{one-to-many} selection), while the right \emph{Reference} panel displays the original image without overlays for visual verification. For each sample, annotators provide point prompts, a referring description with the target term, and a reasoning description without the target term, enabling unified collection of simple and complex instruction-grounding annotations.}
    \vspace{-.2cm}
	\label{fig:tool}
\end{figure*}

\begin{figure*}
	\centering
    \includegraphics[width=1\linewidth]{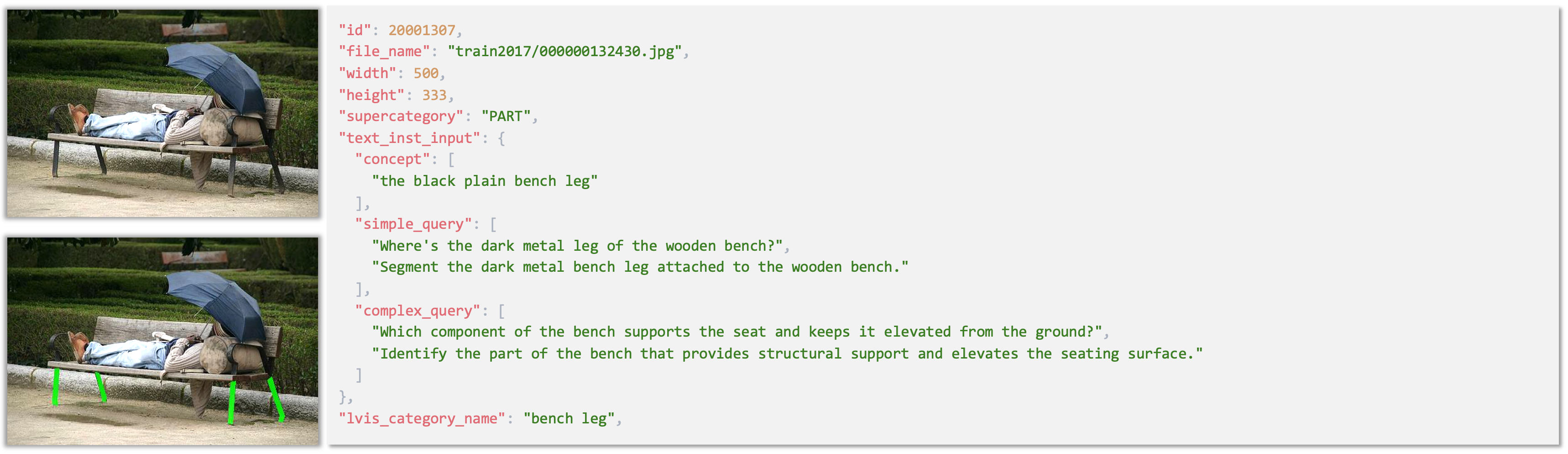}
    \vspace{0.01cm}

	\includegraphics[width=1\linewidth]{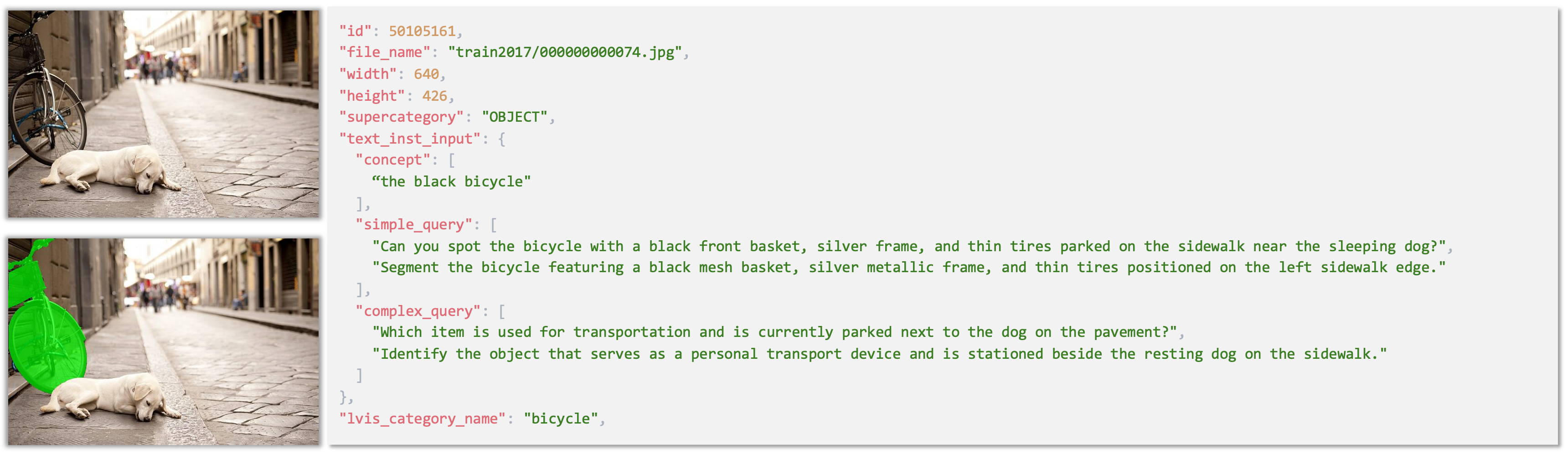}
    \vspace{0.01cm}

    \includegraphics[width=1\linewidth]{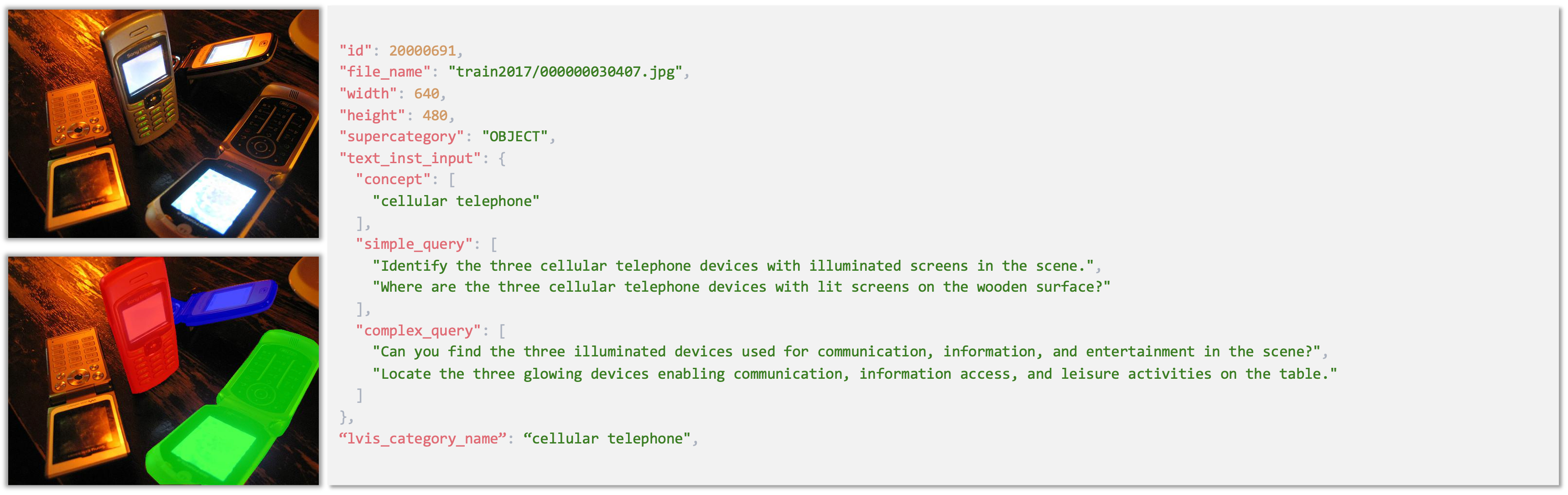}
    \vspace{0.01cm}

    \includegraphics[width=1\linewidth]{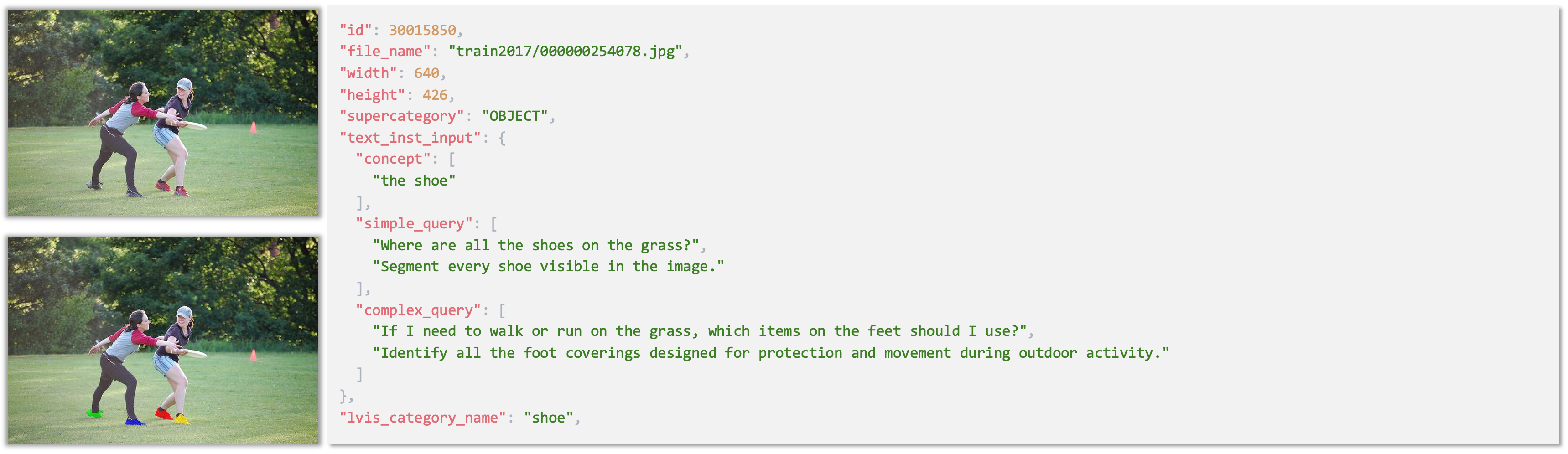}
    \caption{\textbf{Examples from the proposed HMPL-Instruct dataset.} Each sample is represented as a structured JSON entry, incorporating a hierarchical instruction taxonomy with \textit{concept-}, \textit{simple-}, and \textit{complex-level queries} for both object- and part-level segmentation. Target instances are visualized with colored masks. The first example illustrates a part-level segmentation case, while the remaining examples demonstrate object-level scenarios with \textit{one-to-one}, \textit{one-to-many}, and \textit{one-to-all} instruction-instance mappings.}
	\label{fig:data1}
\end{figure*}

\section{Data Construction Details}
\label{Appx:dataset}

\subsection{Human-in-the-Loop Data Construction}

As described in Sec.~\ref{sec:data}, we develop a \textit{human-in-the-loop data engine}. Fig.~\ref{fig:engine} illustrates its overall workflow in detail. This engine yields rich instruction-mask pairs that primarily support \textit{one-to-one} and \textit{one-to-all} grounding scenarios. For the more challenging \textit{one-to-many} setting, where an instruction refers to a \textit{subset} of visually similar instances, we develop a web-based dual-panel instruction annotation tool to facilitate high-quality instruction-grounding annotations, especially for subset-level (\textit{one-to-many}) scenarios. The tool supports both \textit{Annotation} and \textit{Inspection} workflows.

\noindent\textbf{Design and Rationale.} As shown in Fig.~\ref{fig:tool}, the interface displays the same sample in two synchronized panels. The left panel (\textit{Annotate}) is interactive and is used to place point prompts on an image with mask overlays. Importantly, in our setting, point prompts are used as a subset selection signal: when multiple instances of the same concept exist in a scene (\eg, ten persons), annotators may write an instruction that refers to only a subset (\eg, five of them). The clicks explicitly indicate which instances are intended by the instruction. Since the source dataset provides instance masks, we subsequently assign each click to its corresponding mask region (via a point-in-mask check), thereby determining the selected target subset in a reliable and scalable manner. The right panel (\textit{Reference}) is non-interactive and shows the original image without overlays, enabling annotators and inspectors to verify fine visual cues (\eg, color, material, and subtle boundaries) that might be partially occluded by mask visualizations.

\noindent\textbf{Annotation Fields and Their Roles.} For each image, the tool records three complementary supervision signals: i) \textit{Point prompts.} Annotators click on the left panel to add or delete points. All coordinates are stored in the original image resolution to remain invariant to display scaling. These points serve as the instance selection interface for subset-level grounding and enable unambiguous mapping from an instruction to the corresponding mask subset. ii) \textit{Referring description (with target term).} Annotators write a concise referring instruction that explicitly contains the target term (e.g., ``bucket''), serving as a direct grounding cue for simple instructions. iii) \textit{Reasoning description (without target term).} Annotators provide a semantically equivalent instruction that intentionally excludes the target term, requiring implicit identification through functional, relational, or contextual reasoning. This supports complex instruction evaluation and training, where the model must infer targets without explicit naming. Together, these components form a structured annotation unit that supports both explicit referring and implicit reasoning under a unified interface.

\noindent\textbf{Usability features for annotation and inspection.} To support efficient annotation and iterative review, the tool provides: i) a full-screen mode for the left panel with 1$\times$/2$\times$/3$\times$ zoom and Shift+drag panning for precise point placement; ii) auto-save to local storage after any update to prevent annotation loss; iii) export/import utilities, including per-sample JSON export, saving all annotations into a single \texttt{annotation.json}, and loading existing JSON files for resuming and secondary inspection; and iv) efficient navigation via previous/next browsing and jump-by-filename for targeted verification and error correction.

\noindent\textbf{Annotation and Inspection Protocol.} We adopt a two-stage human workflow. During \textit{Annotation}, annotators i) place points to indicate the intended target instances (or subsets), ii) write a referring description with the target term, and iii) write a reasoning description without the target term while following predefined constraints (\eg, complex instructions must not reveal the target NP). During \textit{Inspection}, reviewers use the right reference panel to verify the faithfulness of the textual descriptions to the original visual evidence and check the consistency between selected points and intended targets. Samples with unavoidable referential ambiguity (particularly in subset selection) are flagged and removed.

Overall, the dual-panel tool enables collection of point-assisted subset grounding together with paired referring and reasoning descriptions, while supporting robust human verification via a clear separation between interactive annotation and non-interactive reference views.

\begin{figure*}
	\centering
    \vspace{-0.2cm}
    \includegraphics[width=0.95\linewidth]{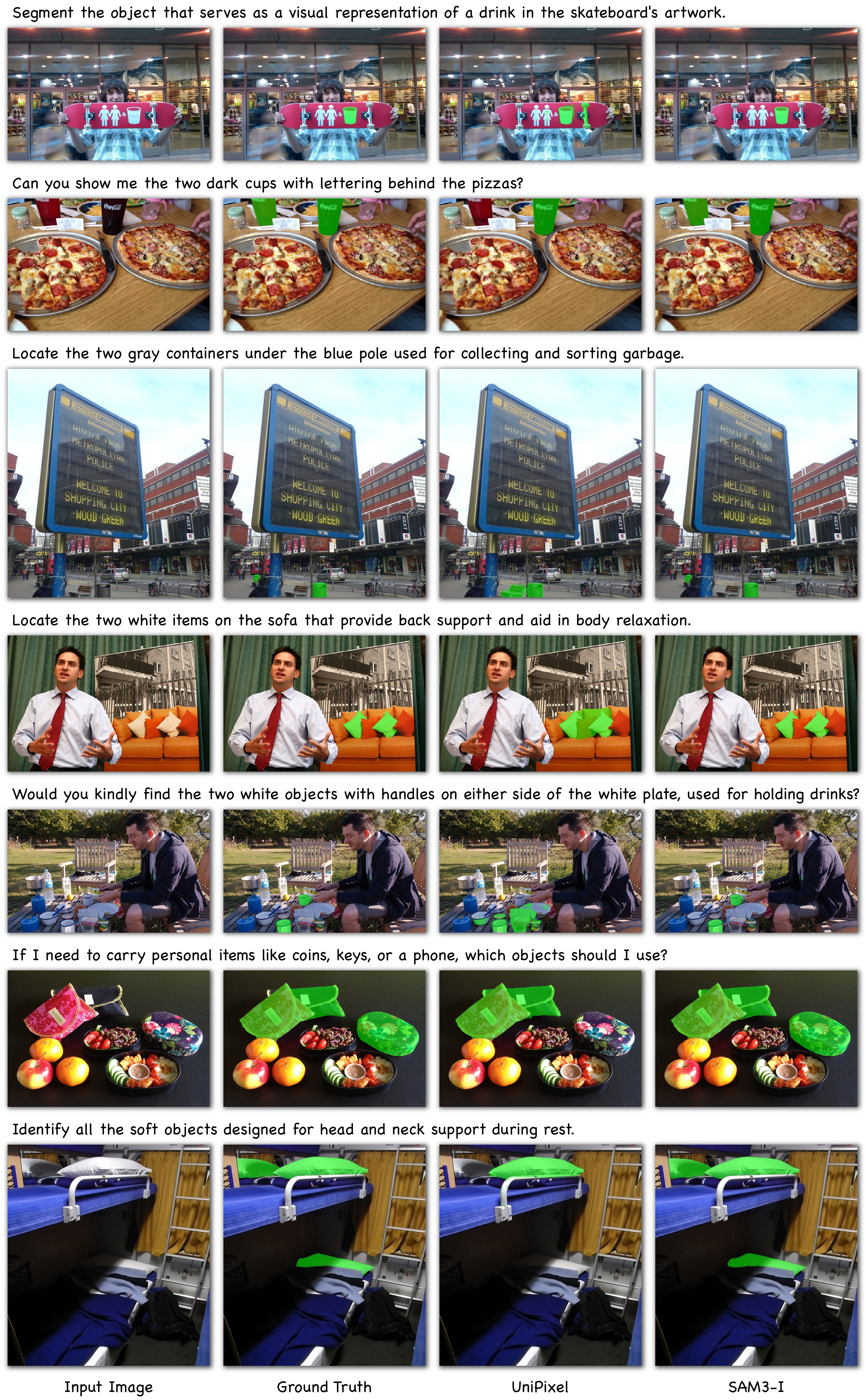}
    \vspace{-0.3cm}
    \caption{\textbf{Visualization results on representative instruction-guided segmentation approaches.}}
	\label{fig:all}
   
\end{figure*}

\section{More Experimental Results}
\label{Appx:results}

\subsection{Hyperparameter Analysis} 
\label{sec:hyper}

We further analyze the sensitivity of SAM3-I to the key hyper-parameters $\rho$ and $m$, which respectively control the weighting factor in the instruction contrastive loss and the margin in the parent-rank concept anchoring loss. 

In Tab.~\ref{tab:abl-rou}, we observe that setting $\rho=0.5$ yields the best overall performance. When $\rho$ is too small, negative samples from the same concept category are overly down-weighted, weakening intra-class discrimination and causing instruction embeddings to collapse toward coarse concept-level representations. Conversely, a large $\rho$ enforces excessive repulsion among same-concept instances, which harms semantic consistency across different instruction expressions and destabilizes training. Setting 
$\rho=0.5$ provides a balanced trade-off between semantic invariance and instance-level discriminability, leading to optimal performance.

In Tab.~\ref{tab:abl-m}, we also examine the effect of the margin parameter $m$ in $\mathcal{L}_{\text{anchor}}$. We find that larger margins result in noticeable performance degradation, indicating overly restrictive semantic constraints.
A moderate margin of $m=0.2$ achieves the best performance.
These results suggest that the proposed parent-rank concept anchoring loss effectively regularizes instruction embeddings to remain aligned with their parent concept semantics, while preserving flexibility for subset-level and instance-specific instruction understanding.

\begin{table}[htbp]
\centering
\small
\caption{\textbf{Ablation analysis on hyper-parameter $\rho$.} The performance is the average of simple and complex scenarios.}
\begin{tabular}{c c c c} 
& {$\rho=0.3$} & {$\rho=0.5$} & {$\rho=0.7$} \\
\toprule
 {\tt\small gIoU} & 54.1 & {54.4} & 52.8 \\
\end{tabular}
\label{tab:abl-rou}
\end{table}

\begin{table}[htbp]
\centering
\small
\caption{\textbf{Ablation analysis on hyper-parameter $m$.} The performance is the average of simple and complex scenarios.}
\begin{tabular}{c c c c} 
& {$m=0.1$} & {$m=0.2$} & {$m=0.3$}  \\
\toprule
{\tt\small gIoU} & 54.2   & {54.4} & 53.7 \\
\end{tabular}
\label{tab:abl-m}
\end{table}

\begin{table}[htbp]
  \centering
  \small
  \caption{\textbf{Quantitative results on the HMPL-Instruct dataset under both object and part levels.}}
  \vspace{-.2cm}
\resizebox{!}{1.3 cm}{\begin{tabular}{l m{0.55cm}<{\centering} m{0.55cm}<{\centering} m{0.55cm}<{\centering}m{0.55cm}<{\centering}} 
& \multicolumn{2}{c}{\textit{Object Level}}  & \multicolumn{2}{c}{\textit{Part Level}} \\
\cmidrule(lr){2-3}\cmidrule(lr){4-5}
{Model} 
& {\tt\small gIoU} & {\tt\small P@50} 
& {\tt\small gIoU} & {\tt\small P@50} \\
\toprule
LISA~\citep{lai2024lisa}            & 31.5 & 26.7 & 13.9 & 8.2 \\
UniPixel~\cite{liu2025unipixel}     & 42.5 & 45.4 & 23.4 & 17.5 \\
SAM3 Agent              & 38.7 & 42.4 & 20.7 & 19.8 \\
{\textbf{SAM3-I}}               & {66.6} & {72.2} & {47.2} & {50.0} \\
\end{tabular}}
\label{tab:ourPart}%
\end{table}

\subsection{More Quantitative and Qualitative Results} 
In Tables~\ref{tab:ourPart}, we report additional quantitative results on the HMPL-Instruct dataset at both the object and part levels. These supplementary results provide a more comprehensive view of model behavior across different target granularities.

We further present additional visualizations to illustrate the characteristics of the HMPL-Instruct dataset, as shown in Fig.~\ref{fig:data1}. The examples span object- and part-level scenarios, covering multiple instruction hierarchies (\textit{concept}, \textit{simple}, and \textit{complex}) as well as diverse instruction–instance mappings (\textit{one-to-one}, \textit{one-to-many}, and \textit{one-to-all}). In addition, Fig.~\ref{fig:all} visualizes annotated targets alongside model predictions, offering qualitative insights into how the most recent UniPixel~\cite{liu2025unipixel} and our SAM3-I respond to varying \textit{one-to-many} instructions reasoning. {Importantly, in these cases, the target cannot be determined from a concept category alone and must be inferred from higher-level semantics. For example, in the fifth raw of Fig.~\ref{fig:all}, the instruction refers to “the two white objects with handles on either side of the white plate, used for holding drinks”. Target identification relies on a combination of attributes, quantity constraints, spatial relations, and functional cues, rather than a concept label such as cup. These results demonstrate that SAM3-I goes beyond latent concept bias and genuinely learns to ground instructions through relational and functional reasoning, particularly in complex, NP-free scenarios.}

\section{Implementation Details}
\label{Appx:imple}

For data engine construction, we adopt two separate Qwen3-VL-8B (Instruct) models~\citep{Qwen3VL} to perform the annotation–inspection loop described in Sec.~\ref{sec:data}. The entire process is executed on 32 NVIDIA H100 GPUs to support large-scale instruction generation and verification. Subsequent model training is conducted with a batch size of 32 on 16 NVIDIA H100 GPUs. The hyperparameters $\rho$ and $m$ are set to 0.5 and 0.2, respectively. The temperature parameter $\tau$ in Eq.~(\ref{equ:loss_stoc}) is set to 0.1 following~\citep{chen2020simple}. Unless otherwise specified, all other hyper-parameters follow the original SAM3 settings without modification. For the SAM3 agent, we use Qwen3-VL-8B as the external MLLM agent and set the maximum number of iterative checking rounds to 3. \textit{Source code and HMPL-Instruct dataset are available at \url{https://github.com/debby-0527/SAM3-I} to support future research on instruction-based segmentation.}

{For the cascaded adapters, they are inserted as residual-parallel branches to the residual architectures in SAM3's text encoder and detector. Specifically, in the SAM3 text encoder, adapters are inserted into each Transformer block as parallel residual modules alongside the feed-forward network, enabling instruction-aware modulation of textual features without modifying the original representations. In the SAM3 detector, adapters are inserted in two locations. First, they are added to the transformer blocks in the multimodal decoder using the same residual-parallel design as in the text encoder. Second, adapters are inserted into the pixel decoder as parallel residual modules along the feature pyramid network connections, allowing instruction semantics to effectively propagate into mask prediction. In detail, each adapter adopts a bottleneck structure with a down-projection to dimension 64, followed by a GELU activation and an up-projection back to the original dimension. The internal MHSA operates at feature dimension 1024 with 4 attention heads. The total number of trainable parameters introduced by the adapters is approximately 311.9M (0.3B). All backbone parameters remain frozen during training.}

\section{Artifact \& Ethics Statement}
\label{Appx:ethics}
\subsection{External Artifacts and Attribution}
This work builds upon several publicly available datasets and models. We use PACO-LVIS~\citep{paco} as the base dataset for constructing HMPL-Instruct, and SAM3~\citep{sam3} as the segmentation foundation. For agent-based baselines and instruction verification, we use Qwen-VL models~\citep{Qwen3VL}. All external artifacts are properly cited in the main paper.

\subsection{License and Intended Use}
All external datasets and models used in this work are released for research purposes, and our usage complies with their original licenses and intended scope. HMPL-Instruct dataset is derived solely from publicly available annotations and images and is intended for academic research on instruction-following segmentation. The dataset is not designed for commercial deployment.

\subsection{Privacy}
HMPL-Instruct does not introduce new personally identifiable information. All images originate from existing public datasets and retain their original privacy and ethical characteristics. No additional personal data is collected, inferred, or annotated in the construction process.

\subsection{Dataset Statistics and Documentation}
We provide comprehensive dataset statistics in Sec.~\ref{sec:data} and Tab.~\ref{tab:dataset}, including the number of samples, instruction categories (concept, simple, complex), target granularities (one-to-one, one-to-many, one-to-all), and object- and part-level annotations. Detailed descriptions of the data construction pipeline and annotation process are also provided.

\end{document}